\documentclass[11pt,a4paper]{article}
\usepackage{acl2015}
\usepackage{amsfonts}
\usepackage{times}
\usepackage{url}
\usepackage{latexsym}
 \usepackage[round]{natbib} 
\usepackage{tabularx}
\usepackage{subcaption}
\usepackage{graphicx}
\usepackage{mathtools}
\usepackage{caption}
\usepackage{subcaption}
\usepackage{algorithm}
\usepackage[font={small}]{caption}
\usepackage{mathtools}

\usepackage[usenames,dvipsnames]{color}
\definecolor{myblue}{rgb}{0,0.1,0.6}
\definecolor{mygreen}{rgb}{0,0.3,0.1}
\usepackage[colorlinks=true,linkcolor=black,citecolor=mygreen,urlcolor=myblue]{hyperref}

\newenvironment{itemizesquish}{\begin{list}{\labelitemi}{\setlength{\itemsep}{0em}\setlength{\labelwidth}{2em}\setlength{\leftmargin}{\labelwidth}\addtolength{\leftmargin}{\labelsep}}}{\end{list}}
\newcommand{\ignore}[1]{}
\newcommand{\transpose}{^\mathsf{T}}

\usepackage{amsthm}
\theoremstyle{definition}
\newtheorem{defn}{Definition} 

\renewcommand{\baselinestretch}{0.962}

\title{
Posterior calibration and exploratory analysis
for natural language processing models
}

\author{
  Khanh Nguyen \\
  Department of Computer Science \\
  University of Maryland, College Park \\
  College Park, MD 20742 \\
  {\tt kxnguyen@cs.umd.edu} \\\And
  Brendan O'Connor \\
  College of Information and Computer Sciences \\
  University of Massachusetts, Amherst \\
  Amherst, MA, 01003 \\
  {\tt brenocon@cs.umass.edu} \\  
}

\date{}

\begin{document}

\maketitle

\begin{abstract}
Many models in natural language processing
define probabilistic distributions over linguistic structures.
We argue that
(1) the quality of a model's posterior distribution
can and should be directly evaluated,
as to whether  probabilities correspond
to empirical frequencies;
and (2) NLP uncertainty
can be projected not only to pipeline components,
but also to exploratory data analysis,
telling a user when to trust and not trust the NLP analysis.
We present a method to analyze calibration,
and apply it to compare the miscalibration
of several commonly used models.
We also contribute a coreference sampling algorithm
that can create confidence intervals for a political event extraction task.\footnote{This is the extended version of a paper published in \emph{Proceedings of EMNLP 2015}.  This version includes acknowledgments and an appendix.  For all materials, see: \url{http://brenocon.com/nlpcalib/}}
\end{abstract}

\section{Introduction}

Natural language processing systems are imperfect.
Decades of research have yielded analyzers that mis-identify named entities,
mis-attach syntactic relations,
and mis-recognize
noun phrase coreference
anywhere from
10-40\% of the time. 
But these systems are accurate enough so that their outputs
can be used as soft, if noisy, indicators of language meaning for use in downstream analysis,
such as systems that perform
question answering, machine translation, event extraction, and narrative analysis
\citep{McCord2012Parsing,Gimpel2008MT,Miwa2010Dependencies,Bamman2013Personas}.

To understand the performance of an analyzer,
researchers and practitioners typically
measure the accuracy of individual labels or edges
among a single predicted output structure $\mathbf{y}$,
such as a most-probable tagging or entity clustering
$\arg\max_{\mathbf{y}} P(\mathbf{y}|x)$
(conditional on text data $x$).

But a probabilistic model gives a probability distribution over many other output structures
that have smaller predicted probabilities; a line of work has sought to control cascading pipeline errors by passing on multiple structures from earlier stages of analysis,
by propagating 
prediction uncertainty
through multiple samples \citep{Finkel2006Pipeline},
$K$-best lists \citep{venugopal2008pipelines,toutanova2008global},
or explicitly diverse lists \citep{gimpel2013diversity};
often the goal is to marginalize over structures to calculate and minimize
an expected loss function,
as in minimum Bayes risk decoding \citep{Goodman1996MBR,Kumar200MBR},
or to perform joint inference between early and later stages of NLP analysis
(e.g.~\citealp{singh2013joint,durrett2014joint}).

These approaches should work better when
the posterior probabilities of the predicted linguistic structures
reflect actual probabilities of the structures or aspects of the structures.
For example, say a model is overconfident: it places too much probability mass in the top prediction, and not enough in the rest.
Then there will be little benefit to using the lower probability structures, since
in the training or inference objectives
they will be incorrectly outweighed by the top prediction
(or in a sampling approach, they will be systematically undersampled and thus have too-low frequencies).
If we only evaluate models based on their top predictions 
or on downstream tasks, it is difficult to diagnose this issue.

Instead, we propose to directly evaluate the \emph{calibration}
of a model's posterior prediction distribution.
A perfectly calibrated model knows how often it's right or wrong;
when it predicts an event with 80\% confidence,
the event empirically turns out to be true 80\% of the time.
While perfect accuracy for NLP models remains an unsolved challenge,
perfect calibration is a more achievable goal,
since a model that has imperfect accuracy could, in principle, be perfectly calibrated.
In this paper, we develop a method to empirically analyze calibration that is appropriate for NLP models (\S\ref{s:empircalib})
and use it to analyze common generative and discriminative models 
for tagging and classification
(\S\ref{s:classtag}).

Furthermore, if a model's probabilities are meaningful,
that would justify using its probability distributions for any downstream purpose,
including exploratory analysis on unlabeled data.
In \S\ref{s:entevt} we introduce a representative corpus exploration problem, identifying temporal event trends in international politics, with a method
that is dependent on coreference resolution.
We develop a coreference sampling algorithm
(\S\ref{s:corefsampling}) which projects uncertainty into
the event extraction, inducing a posterior distribution over event frequencies.
Sometimes the event trends have very high posterior variance 
(large confidence intervals),\footnote{We use the terms \emph{confidence interval} and \emph{credible interval} interchangeably in this work; the latter term is debatably more correct, though less widely familiar.}
reflecting when the NLP system genuinely does not know the correct semantic extraction.
This highlights an important use of a calibrated model: being able to tell a user when the model's predictions are likely to be incorrect, or at least, not giving a user a false sense of certainty from an erroneous NLP analysis.


\section{Definition of calibration}
\label{s:defcalib}
Consider a binary probabilistic prediction problem, which consists of binary labels and probabilistic predictions for them.
Each instance has a \emph{ground-truth label} $y \in \{0, 1\}$, which is used for evaluation.  The prediction problem is to generate a \emph{predicted probability} or \emph{prediction strength} $q \in [0,1]$.  Typically, we use some form of a probabilistic model to accomplish this task, where $q$ represents the model's posterior probability\footnote{Whether $q$ comes from a Bayesian posterior or not
is irrelevant to the analysis in this section.
All that matters is that predictions are numbers $q \in [0,1]$.}
of the instance having a positive label ($y=1$).

Let $S = \{(q_1, y_1), (q_2, y_2), \cdots\, (q_N, y_N)\}$ be the set of prediction-label pairs produced by the model. Many metrics assess the overall quality of how well the predicted probabilities match the data, such as the familiar cross entropy (negative average log-likelihood),
\[ L_\ell(\vec{y},\vec{q}) = \frac{1}{N} \sum_i y_i \log \frac{1}{q_i} + (1-y_i) \log \frac{1}{1-q_i}\]
or mean squared error, also known as the \emph{Brier score} when $y$ is binary \citep{Brier1950},
\[ L_2(\vec{y},\vec{q}) = \frac{1}{N} \sum_i (y_i-q_i)^2    \vspace{-0.05in}\]
Both tend to attain better (lower) values when $q$ is near 1 when $y=1$, and near 0 when $y=0$; and they achieve a perfect value of 0 when all $q_i=y_i$.\footnote{These two loss functions are instances of \emph{proper scoring rules} \citep{Gneiting2007Scoring,Brocker2009Decomposition}.
}

Let $\mathbb{P}(y,q)$ be the joint empirical distribution over labels and predictions.  Under this notation, $L_2 = \mathbb{E}_{q,y}[y-q]^2$. Consider the factorization
\[\mathbb{P}(y,q)=\mathbb{P}(y \mid q)\ \mathbb{P}(q)\]
where $\mathbb{P}(y \mid q)$ denotes the label empirical frequency, conditional on a prediction strength \citep{Murphy1987General}.\footnote{
We alternatively refer to this as \emph{label frequency} or \emph{empirical frequency}. The $\mathbb{P}$ probabilities can be thought of as frequencies from the hypothetical population the data and predictions are drawn from.
$\mathbb{P}$ probabilities are, definitionally speaking, completely separate from a probabilistic model that might be used to generate $q$ predictions.}
Applying this factorization to the Brier score
leads to the calibration-refinement decomposition
\citep{degroot1983comparison}, in terms of
expectations with respect to the prediction strength distribution $\mathbb{P}(q)$:
\begin{equation}
L_2 
\ = \ 
\underbrace{\mathbb{E}_q[q-p_q]^2}_{\text{Calibration MSE}}
\ + \ 
\underbrace{\mathbb{E}_q[p_q(1-p_q)]}_{\text{Refinement}}
\label{e:twoterms}
\end{equation}
where we denote $p_q \equiv \mathbb{P}(y=1 \mid q)$ for brevity.  

Here, \emph{calibration} measures to what extent a model's probabilistic predictions match their corresponding empirical frequencies.  Perfect calibration is achieved when $\mathbb{P}(y=1 \mid q) = q$ for all $q$; intuitively, if you aggregate all instances where a model predicted $q$, they should have $y=1$ at $q$ percent of the time.
We define the magnitude of miscalibration using root mean squared error:
\begin{defn}[RMS calibration error]
\[ CalibErr = \sqrt{ \mathbb{E}_q[ q - \mathbb{P}(y=1 \mid q) ]^2 } \]
\end{defn}
\noindent
The second term of Eq~\ref{e:twoterms}
refers to \emph{refinement}, which reflects
to what extent the model is able to separate different labels
(in terms of the conditional Gini entropy $p_q(1-p_q)$).
If the prediction strengths
tend to cluster around 0 or 1, the refinement score tends to be lower.
The calibration-refinement breakdown offers a useful perspective on the accuracy of a model posterior.  This paper focuses on calibration.


There are several other ways to break down squared error, log-likelihood, and other probabilistic scoring rules.\footnote{They all include a notion of calibration corresponding to a Bregman divergence \citep{Brocker2009Decomposition}; for example, cross-entropy can be broken down such that KL divergence is the measure of miscalibration.}
We use the Brier-based calibration error in this work, since unlike cross-entropy it does not  tend toward infinity when near probability 0; we hypothesize this could be an issue since both $p$ and $q$ are subject to estimation error.

\section{Empirical calibration analysis}
\label{s:empircalib}

\begin{figure}[t]
  \begin{minipage}{\linewidth}
\begin{algorithm}[H]
\renewcommand{\baselinestretch}{0.6}
  \small
  \caption{Estimate calibration error using adaptive binning. \label{alg:alg1}}
  ~\\
  \textbf{Input:} A set of $N$ prediction-label pairs $\{(q_1, y_1), (q_2, y_2), \cdots, (q_N, y_N)\}$.  \\

  \textbf{Output:} Calibration error. \\ 

  \textbf{Parameter:} Target bin size $\beta$. \\

  Step 1: Sort pairs by prediction values $q_k$ in ascending order. \\

  Step 2: For each, assign bin label $b_k= \left\lfloor \frac{k - 1}{\beta}\right\rfloor + 1$. \\
  
  Step 3: Define each bin $B_i$ as the set of indices of pairs that have the same bin label.  If the last bin has size less than $\beta$, merge it with the second-to-last bin (if one exists). Let $\{B_1, B_2, \cdots, B_T\}$ be the set of bins.\\

  Step 4: Calculate empirical and predicted probabilities per bin: \vspace{-0.05in}
$$\hat{p}_i = \frac{1}{|B_i|}\sum_{k \in B_i}y_k
\ \ \ \ \ \ \ \text{and}\ \ \ \ \ \ \ 
\hat{q}_i = \frac{1}{|B_i|}\sum_{k \in B_i}q_k$$ 
  Step 5: Calculate the calibration error as the root mean squared error per bin,
  weighted by bin size in case they are not uniformly sized:  \vspace{-0.05in}
$$CalibErr = \sqrt{\frac{1}{N}\sum_{i=1}^T|B_i|(\hat{q}_i - \hat{p}_i)^2}$$
\vspace{-3mm}
\end{algorithm}
\end{minipage}
  \vspace{-0.2in}
\end{figure}

\begin{figure}[t]
  \begin{minipage}{\linewidth} 
  \begin{algorithm}[H]
\renewcommand{\baselinestretch}{0.6}
    \small
    \caption{Estimate calibration error's confidence interval by sampling.     \label{alg:alg2}} 
     ~\\
     \textbf{Input:}  A set of $N$ prediction-label pairs $\{(q_1, y_1), (q_2, y_2), \cdots, (q_N, y_N)\}$.  \\

     \textbf{Output:} Calibration error with a 95\% confidence interval. \\

     \textbf{Parameter:} Number of samples, $S$. \\

     Step 1: Calculate $\{\hat{p}_1, \hat{p}_2, \cdots, \hat{p}_T\}$ from step 4 of Algorithm~\ref{alg:alg1}.\\
     
     Step 2: Draw $S$ samples.  For each $s = 1..S$,
     \begin{itemize}
     \item For each bin $i = 1..T$,
     draw $\hat{p}^{(s)}_i \sim \mathcal{N}\left(\hat{p}_i, \hat{\sigma}^2_i\right)$,
	where $\hat{\sigma}^2_i = \hat{p}_i(1-\hat{p}_i) / |B_i|$.
	If necessary clip to $[0,1]$:
	$\hat{p}^{(s)}_i := \min(1,\max(0, \hat{p}^{(s)}_i))$
      \item Calculate the sample's $CalibErr$ from using the pairs $(\hat{q}_i, \hat{p}^{(s)}_i)$ as per Step 5 of Algorithm~\ref{alg:alg1}.
     \end{itemize}
     Step 3: Calculate the 95\% confidence interval for the calibration error as:

     $$CalibErr_{avg} \pm 1.96\ \hat{s}_{error}$$ where $CalibErr_{avg}$ and $\hat{s}_{error}$ are the mean and the standard deviation, respectively, of the $CalibErr$s calculated from the samples. \\ 
\end{algorithm}
\end{minipage}
  \vspace{-0.2in}
\end{figure}

From a test set of labeled data, we can analyze model calibration both in terms of the calibration error, as well as visualizing the \emph{calibration curve} of label frequency versus predicted strength.
However, computing the label frequencies $\mathbb{P}(y=1|q)$ requires an infinite amount of data. Thus approximation methods are required to perform calibration analysis.

\subsection{Adaptive binning procedure}
\label{s:adapt_bin}

\begin{figure*}[t]
  \begin{minipage}[t]{0.24\textwidth}
    \centering
    \subcaptionbox{\normalsize}
    {\includegraphics[width=\textwidth]{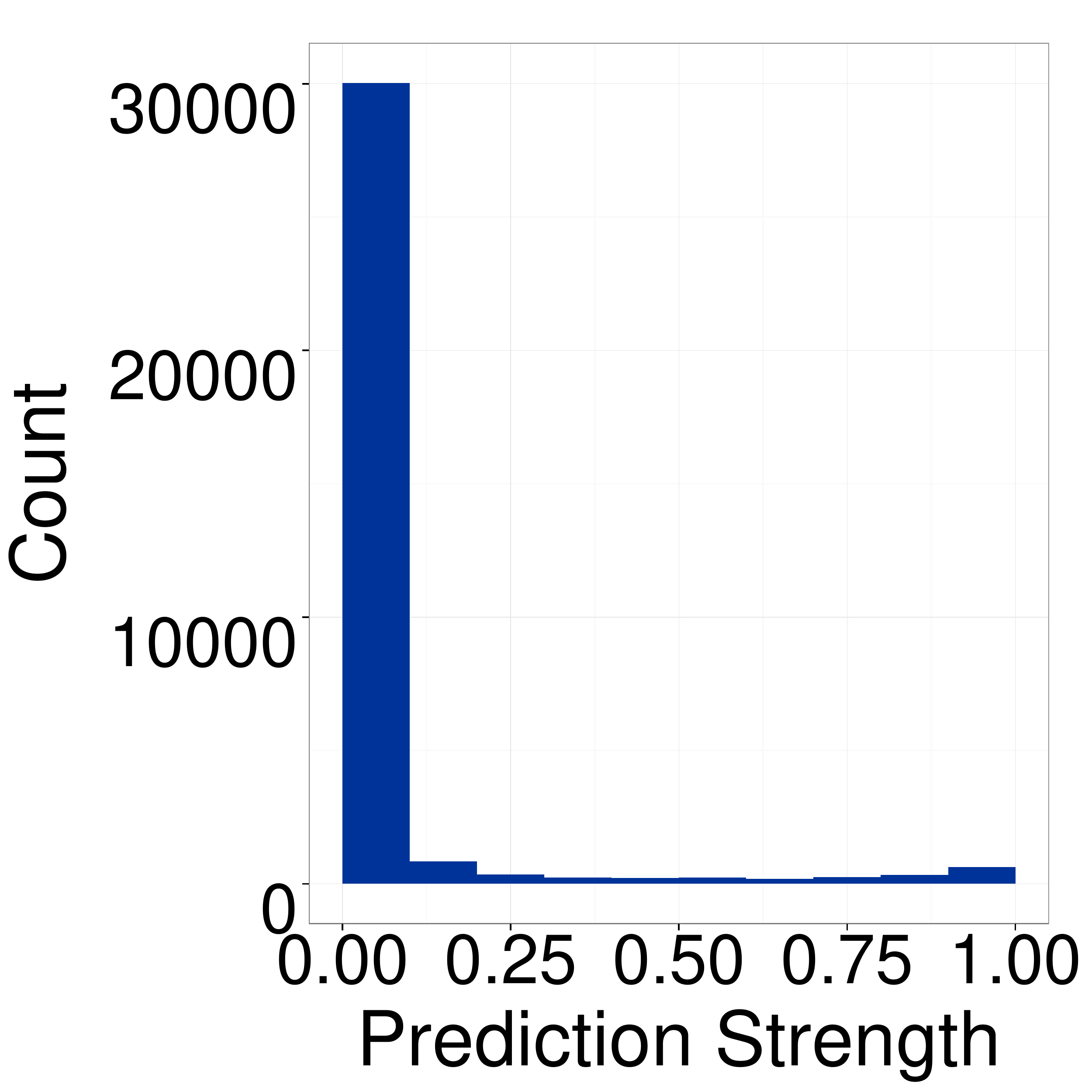}}
  \end{minipage}
  \begin{minipage}[t]{0.24\textwidth}
  \centering
    \subcaptionbox{\normalsize}
    {\includegraphics[width=\textwidth]{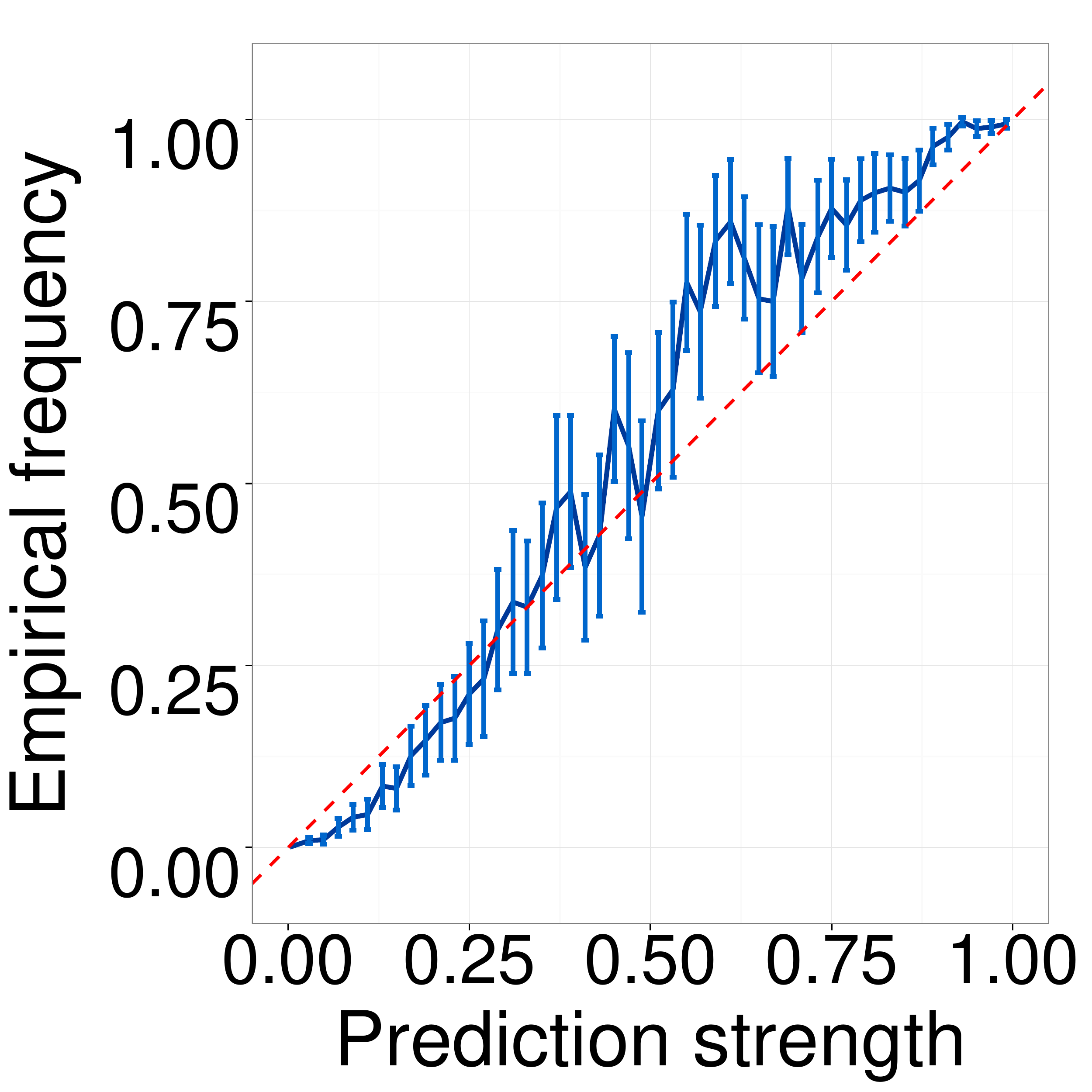}}
  \end{minipage}
  \begin{minipage}[t]{0.24\textwidth}
    \centering
    \subcaptionbox{\normalsize}
    {\includegraphics[width=\textwidth]{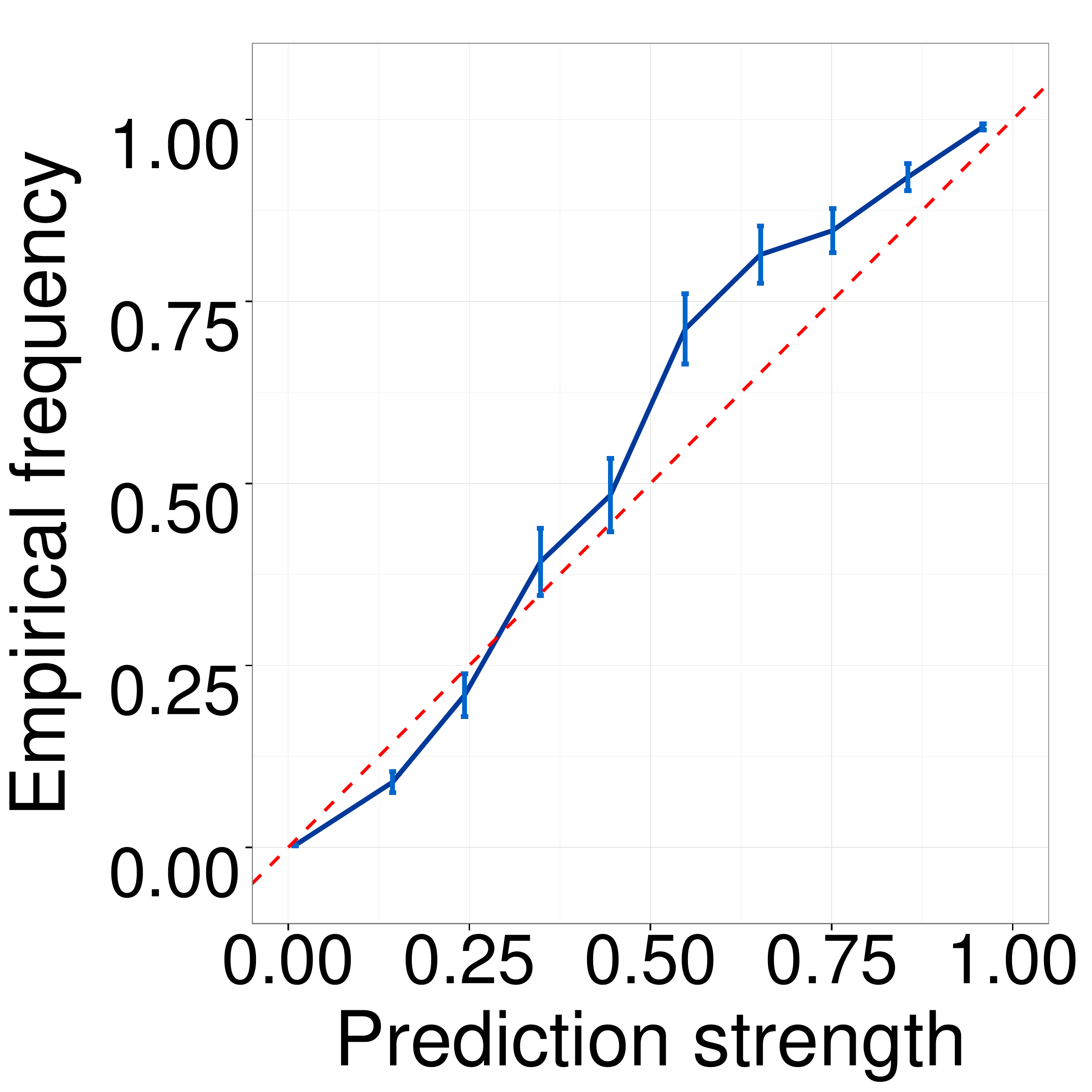}}
  \end{minipage}
  \begin{minipage}[t]{0.24\textwidth}
    \centering
    \subcaptionbox{\normalsize}
    {\includegraphics[width=\textwidth]{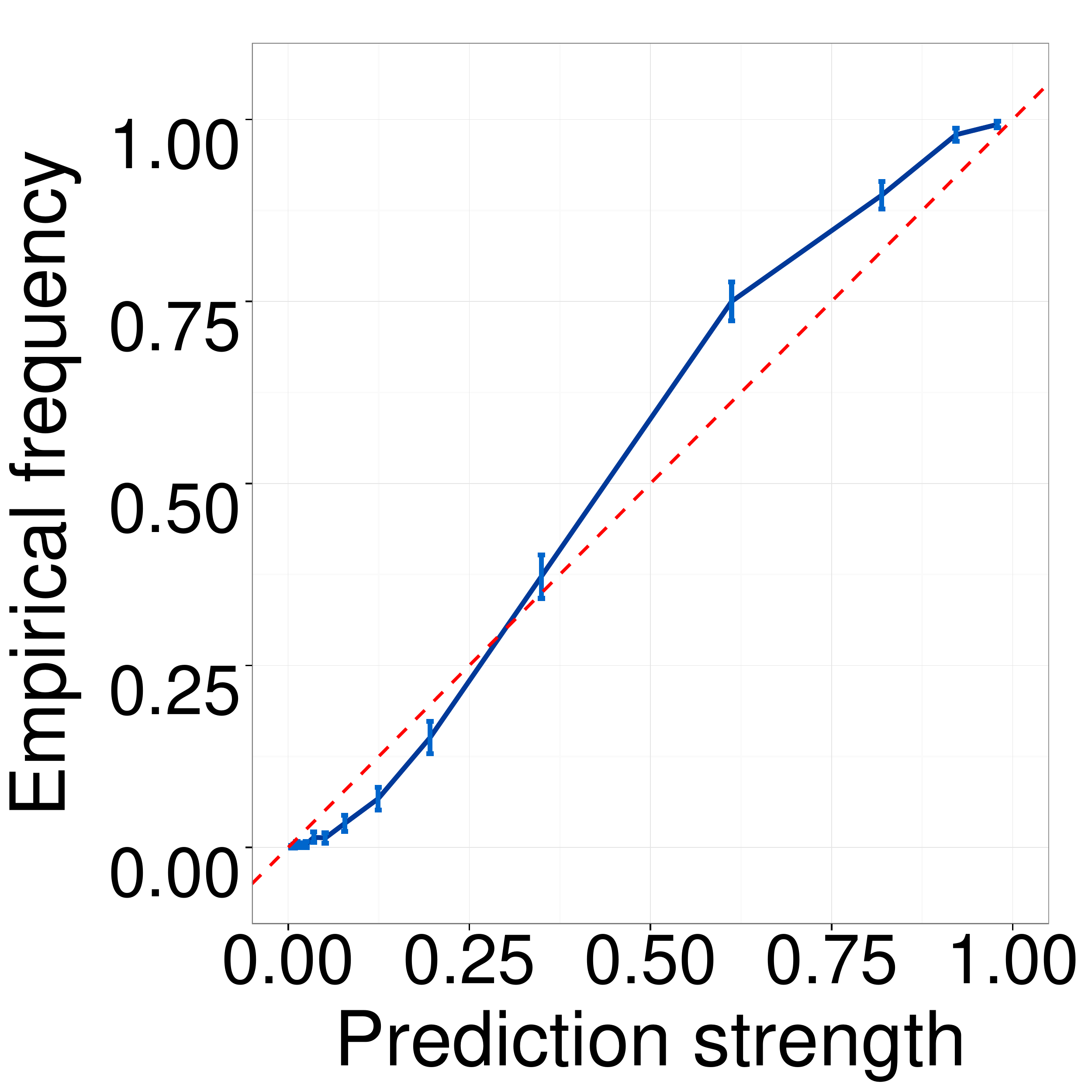}}
  \end{minipage}
    \vspace{-0.1in}
  \caption{(a) A skewed distribution of predictions on whether a word has the NN tag (\S\ref{s:singletags}). Calibration curves produced by equally-spaced binning with bin width equal to 0.02 (b) and 0.1 (c) can have wide confidence intervals. Adaptive binning (with 1000 points in each bin) (d) gives small confidence intervals and also captures the prediction distribution. The confidence intervals are estimated as described in \S\ref{s:adapt_bin}.   \vspace{-0.2in}}
  \label{fig:bin_method}
\end{figure*}

Previous studies that assess calibration in supervised machine learning models
\citep{Niculescu2005Calibration,Bennett2000NB}
calculate label frequencies by dividing the prediction space into deciles or other evenly spaced bins---e.g.~$q \in [0,0.1)$, $q\in[0.1,0.2)$, etc.---and then calculating the empirical label frequency in each bin. This procedure may be thought of as using a form of nonparametric regression (specifically, a regressogram; \citealt{Tukey1961Regressogram}) to estimate the function $f(q)=\mathbb{P}(y=1 \mid q)$
from observed data points.
But models in natural language processing give very skewed distributions of confidence scores $q$ (many are near 0 or 1), so this procedure performs poorly, having much more variable estimates near the middle of the $q$ distribution (Figure \ref{fig:bin_method}).

We propose adaptive binning as an alternative. Instead of dividing the interval $[0, 1]$ into fixed-width bins, adaptive binning defines the bins such that there are an equal number of points in each, after which the same averaging procedure is used.
This method naturally gives wider bins to area with fewer data points (areas that require more smoothing), and ensures that these areas have roughly similar standard errors as those near the boundaries,
since for a bin with $\beta$ number of points and empirical frequency $p$,
the standard error is estimated by $\sqrt{p(1-p)/\beta}$, which is bounded above by $0.5/\sqrt{\beta}$.
Algorithm \ref{alg:alg1} describes the procedure for estimating calibration error using adaptive binning, which can be applied to any probabilistic model that predicts posterior probabilities.

\subsection{Confidence interval estimation}
\label{s:ci}

Especially when the test set is small, estimating calibration error may be subject to error,
due to uncertainty in the label frequency estimates.  
Since how to estimate confidence bands for nonparametric regression
is an unsolved problem \citep{Wasserman2006AllNP},
we resort to a simple method based on the binning.
We construct a binomial normal approximation
for the label frequency estimate in each bin, and simulate from it;
every simulation across all bins is used to construct a calibration error; these simulated calibration errors are collected to construct a normal approximation for the calibration error estimate.
Since we use bin sizes of at least $\beta\geq 200$ in our experiments,
the central limit theorem justifies these approximations.
We report all calibration errors along with their 95\% confidence intervals
calculated by Algorithm \ref{alg:alg2}.\footnote{A major unsolved issue is how to fairly select the bin size.  If it is too large, the curve is oversmoothed and calibration looks better than it should be; if it is too small, calibration looks worse than it should be.  Bandwidth selection and cross-validation techniques may better address this problem in future work.  In the meantime, visualizations of calibration curves
help inform the reader of the resolution of a particular analysis---if the bins are far apart,
the data is sparse, and the specific details of the curve are not known in those regions.}

\subsection{Visualizing calibration}

In order to better understand a model's calibration properties, we plot the pairs
 $(\hat{p}_1, \hat{q}_1), (\hat{p}_2, \hat{q}_2), \cdots, (\hat{p}_T, \hat{q}_T)$
 obtained from the adaptive binning procedure to visualize the \emph{calibration curve} of the model---this visualization is known as a \emph{calibration} or \emph{reliability plot}.
It provides finer grained insight into the calibration behavior in different prediction ranges. A perfectly calibrated curve would coincide with the $y=x$ diagonal line.
When the curve lies above the diagonal, the model is underconfident ($q<p_q$);
and when it is below the diagonal, the model is overconfident ($q>p_q$).

An advantage of plotting a curve estimated from fixed-size bins, instead of fixed-width bins, is that the distribution of the points hints at the refinement aspect of the model's performance.
If the points' positions tend to cluster in the bottom-left and top-right corners, that implies the model is making more refined predictions.

\section{Calibration for classification and tagging models}
\label{s:classtag}

Using the method described in \S\ref{s:empircalib}, we assess the quality of posterior predictions of several classification and tagging models. In all of our experiments, we set the target bin size in Algorithm~\ref{alg:alg1} to be 5,000 and the number of samples in Algorithm~\ref{alg:alg2} to be 10,000. 

\subsection{Naive Bayes and logistic regression}
\label{s:naibay_logreg}

\subsubsection{Introduction}

Previous work on Naive Bayes has found its probabilities to have
calibration issues, in part due to its incorrect conditional independence assumptions \citep{Niculescu2005Calibration,Bennett2000NB,Domingos1997NB}.
Since logistic regression has the same log-linear representational capacity
\citep{Ng2002NBLR} but does not suffer from the independence assumptions,
we select it for comparison,
hypothesizing it may have better calibration.

We analyze a binary classification task of Twitter sentiment analysis from emoticons.
We collect a dataset consisting of tweets identified by the Twitter API as English, collected from 2014 to 2015, with the ``emoticon trick'' \citep{Read2005Emoticons,Lin2012Twitter} to label tweets that contain at least one occurrence of the smiley emoticon ``:)'' as ``happy'' ($y=1$) and others as $y=0$.  The smiley emoticons are deleted in positive examples.  We sampled three sets of tweets (subsampled from the Decahose/Gardenhose stream of public tweets) with Jan-Apr 2014 for training, May-Dec 2014 for development, and Jan-Apr 2015 for testing.
Each set contains $10^5$ tweets, split between an equal number of positive and negative instances.  We use binary features based on unigrams extracted from the \emph{twokenize.py}\footnote{\url{https://github.com/myleott/ark-twokenize-py}} tokenization.
We use the \emph{scikit-learn} \citep{scikit-learn} implementations of Bernoulli Naive Bayes and L2-regularized logistic regression. The models' hyperparameters (Naive Bayes' smoothing paramter and logistic regression's regularization strength) are chosen to maximize the F-1 score on the development set.

\subsubsection{Results}
\label{s:emoticon}

\begin{figure}[t]
\centering
  \subcaptionbox{\normalsize}
  {\includegraphics[width=0.49\linewidth]{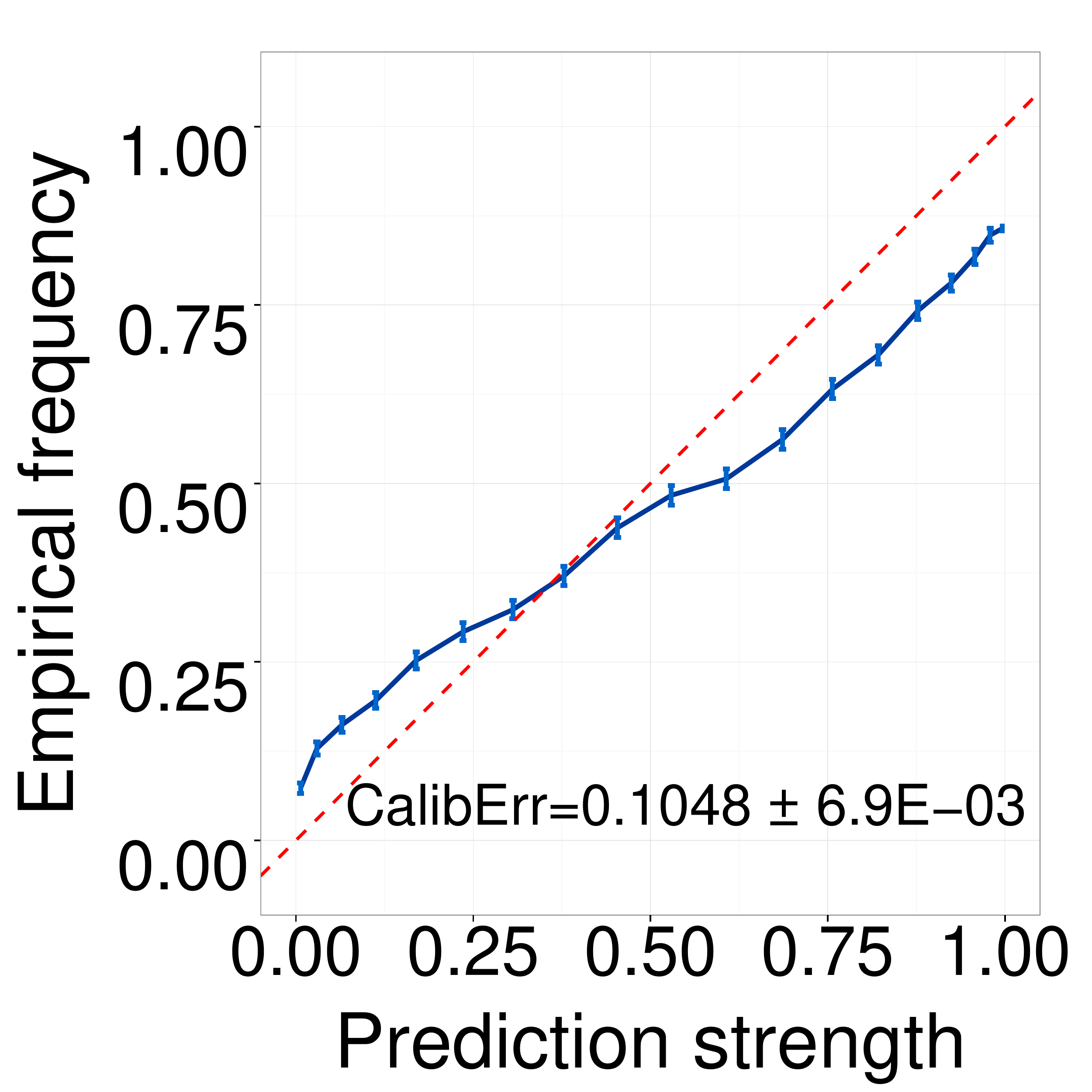}}
  \subcaptionbox{\normalsize}
  {\includegraphics[width=0.49\linewidth]{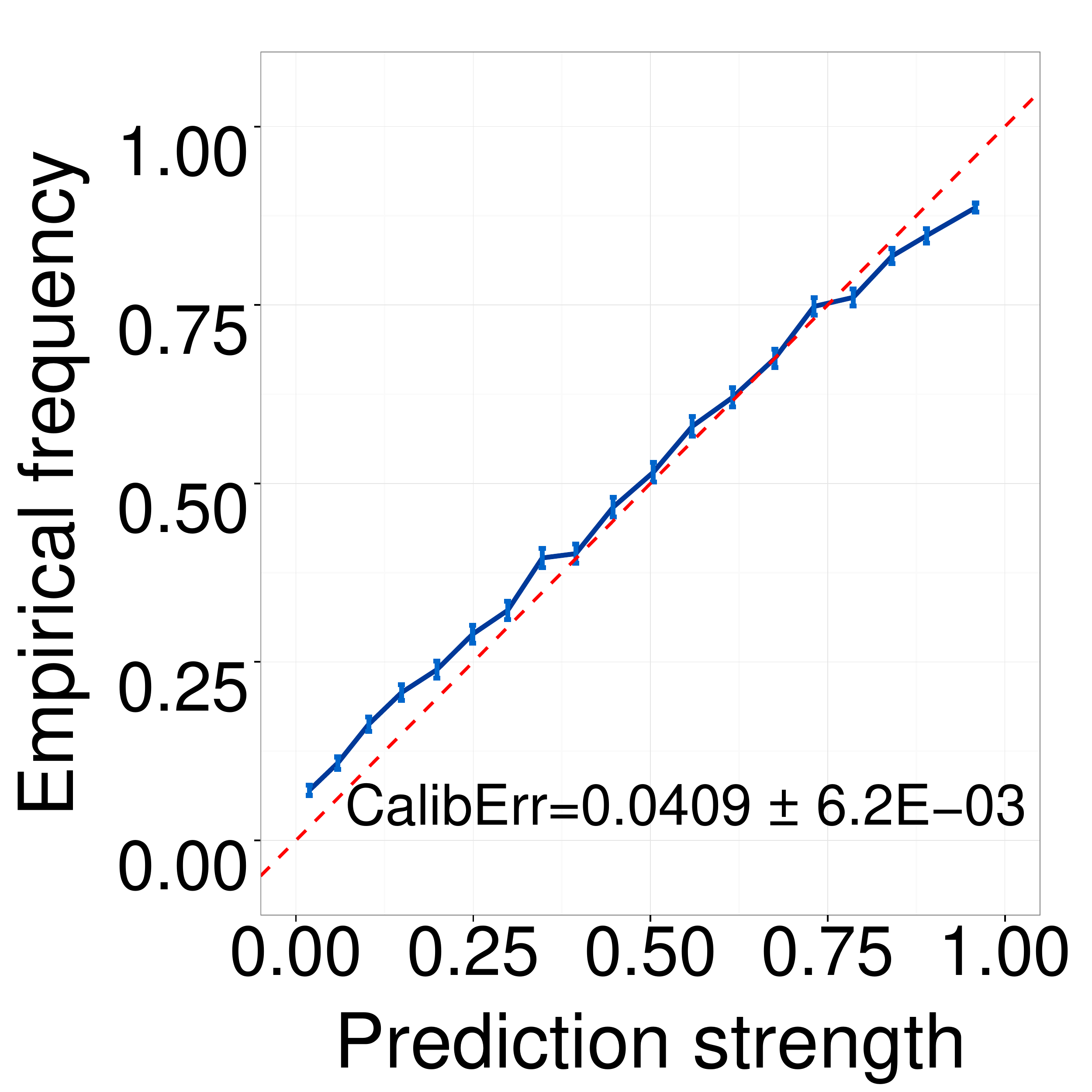}}
  \caption{Calibration curve of (a) Naive Bayes and (b) logistic regression on predicting whether a tweet is a ``happy'' tweet.   \vspace{-0.2in}} 
  \label{fig:emo}
\end{figure}

Naive Bayes attains a slightly higher F-1 score (NB 73.8\% vs.~LR 72.9\%),
but logistic regression has much lower calibration error: less than half as much RMSE
 (NB 0.105 vs.~LR 0.041; Figure \ref{fig:emo}).
Both models have a tendency to be
underconfident in the lower prediction range and overconfident in the higher range,
but the tendency is more pronounced for Naive Bayes.

\subsection{Hidden Markov models and conditional random fields}
\label{s:hmm_crf}

\subsubsection{Introduction}

Hidden Markov models (HMM) and
linear chain conditional random fields (CRF)
are another commonly used pair of analogous generative and discriminative models.
They both define a posterior over tag sequences $P(\mathbf{y}|x)$,
which we apply to part-of-speech tagging.

We can analyze these models in the binary calibration framework (\S\ref{s:defcalib}-\ref{s:empircalib}) by looking at marginal distribution of binary-valued outcomes of parts of the predicted structures.
Specifically, we examine calibration of predicted probabilities of individual tokens' tags (\S\ref{s:singletags}), 
and of pairs of consecutive tags (\S\ref{s:pairtags}).
These quantities are calculated with the forward-backward algorithm.

To prepare a POS tagging dataset, we extract \emph{Wall Street Journal} articles from the English CoNLL-2011 coreference shared task dataset from Ontonotes \citep{pradhan2011conll},
using the CoNLL-2011 splits for training, development and testing.
This results in 11,772 sentences for training, 1,632 for development,
and 1,382 for testing, over a set of 47 possible tags.

We train an HMM with Dirichlet MAP
using one pseudocount for every transition and word emission.
For the CRF, we use the $L_2$-regularized L-BFGS algorithm implemented in \emph{CRFsuite} \citep{CRFsuite}. 
We compare an HMM to a CRF that only uses
basic
transition (tag-tag) and emission (tag-word) features,
so that it does not have an advantage
due to more features.
In order to compare models with similar task performance,
we train the CRF with only 3000 sentences from the training set,
which yields the same accuracy as the HMM (about 88.7\% on the test set).
In each case, the model's hyperparameters (the CRF's $L_2$ regularizer, the HMM's pseudocount) are selected by maximizing accuracy on the development set.

\subsubsection{Predicting single-word tags}
\label{s:singletags}

\begin{figure}[t]
  \centering
  \subcaptionbox{\normalsize}
   {\includegraphics[width=0.49\linewidth]{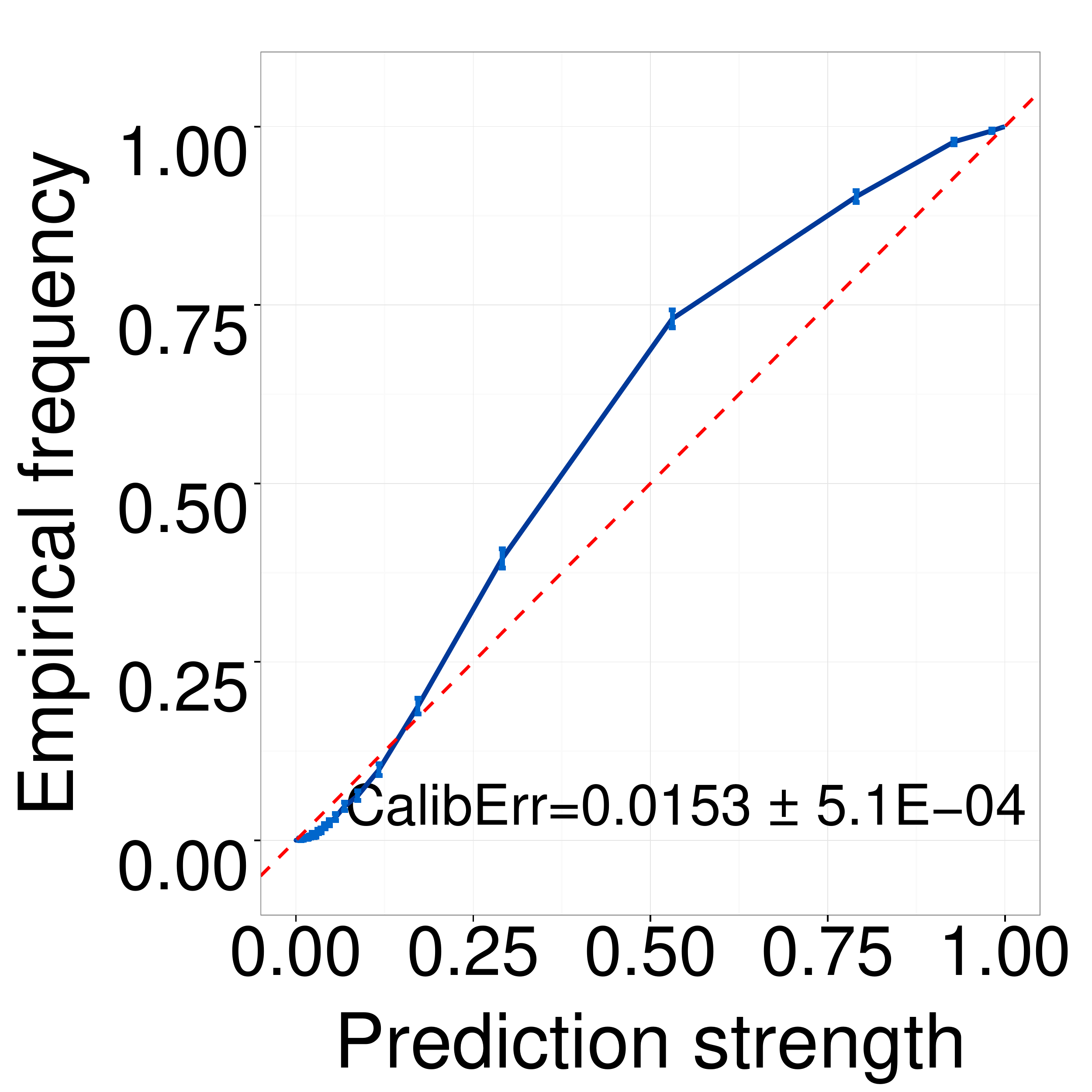}}
   \subcaptionbox{\normalsize}
   {\includegraphics[width=0.49\linewidth]{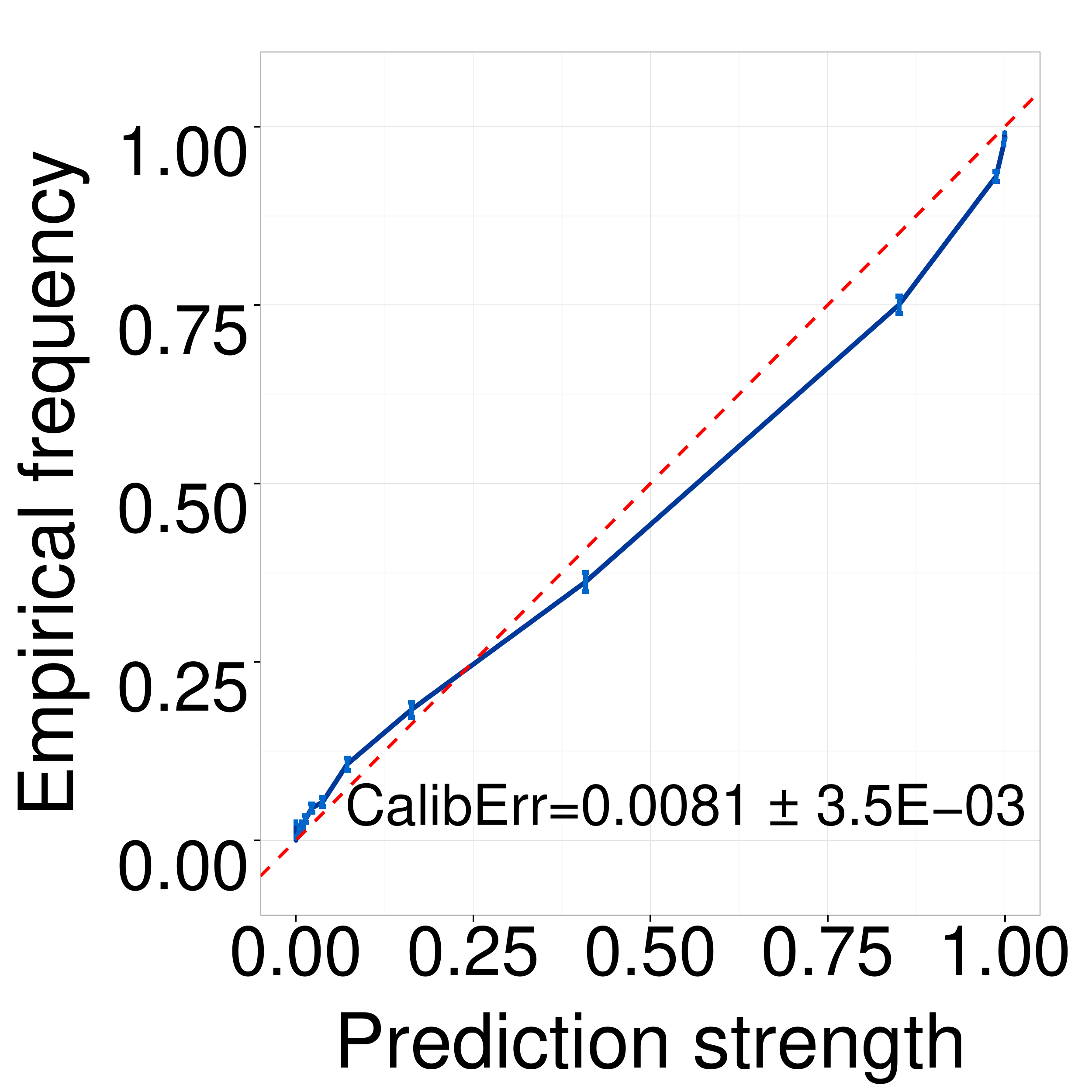}}
  \caption{Calibration curves of (a) HMM, and (b) CRF, on predictions over all POS tags.
  \vspace{-0.2in}}
  \label{fig:pos}
\end{figure}

In this experiment, we measure miscalibration of the two models on predicting tags of single words.  First, for each tag type, we produce a set of 33,306 prediction-label pairs (for every token); we then concatenate them across the tags for calibration analysis.
Figure \ref{fig:pos} shows that the two models exhibit distinct calibration patterns. The HMM tends to be very underconfident whereas the CRF is overconfident, and the CRF has a lower (better) overall calibration error.

We also examine the calibration errors of the individual POS tags (Figure \ref{fig:pos_score}(a)). We find that CRF is significantly better calibrated than HMM in most but not all categories (39 out of 47).
For example, they are about equally calibrated on predicting the NN tag.
The calibration gap between the two models also differs among the tags.  

\subsubsection{Predicting two-consecutive-word tags}
\label{s:pairtags}

\begin{figure}[t]
   \centering
   \vspace{-0.02in}
   \subcaptionbox{\normalsize}
   {\includegraphics[width=3in,clip=true,trim=0 73 0 0]{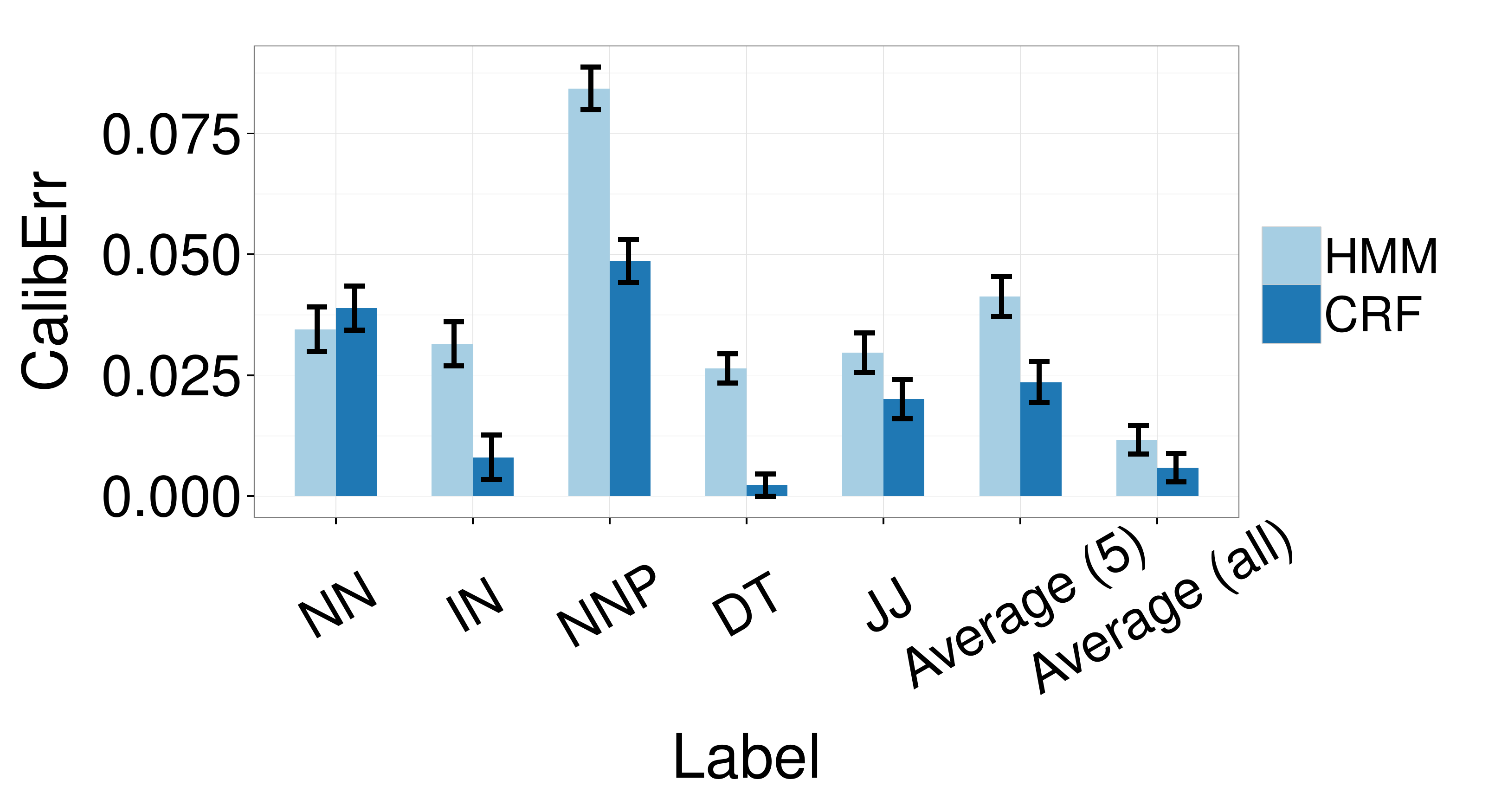}}
   \subcaptionbox{\normalsize}
   {\includegraphics[width=3in,clip=true,trim=0 73 0 25]{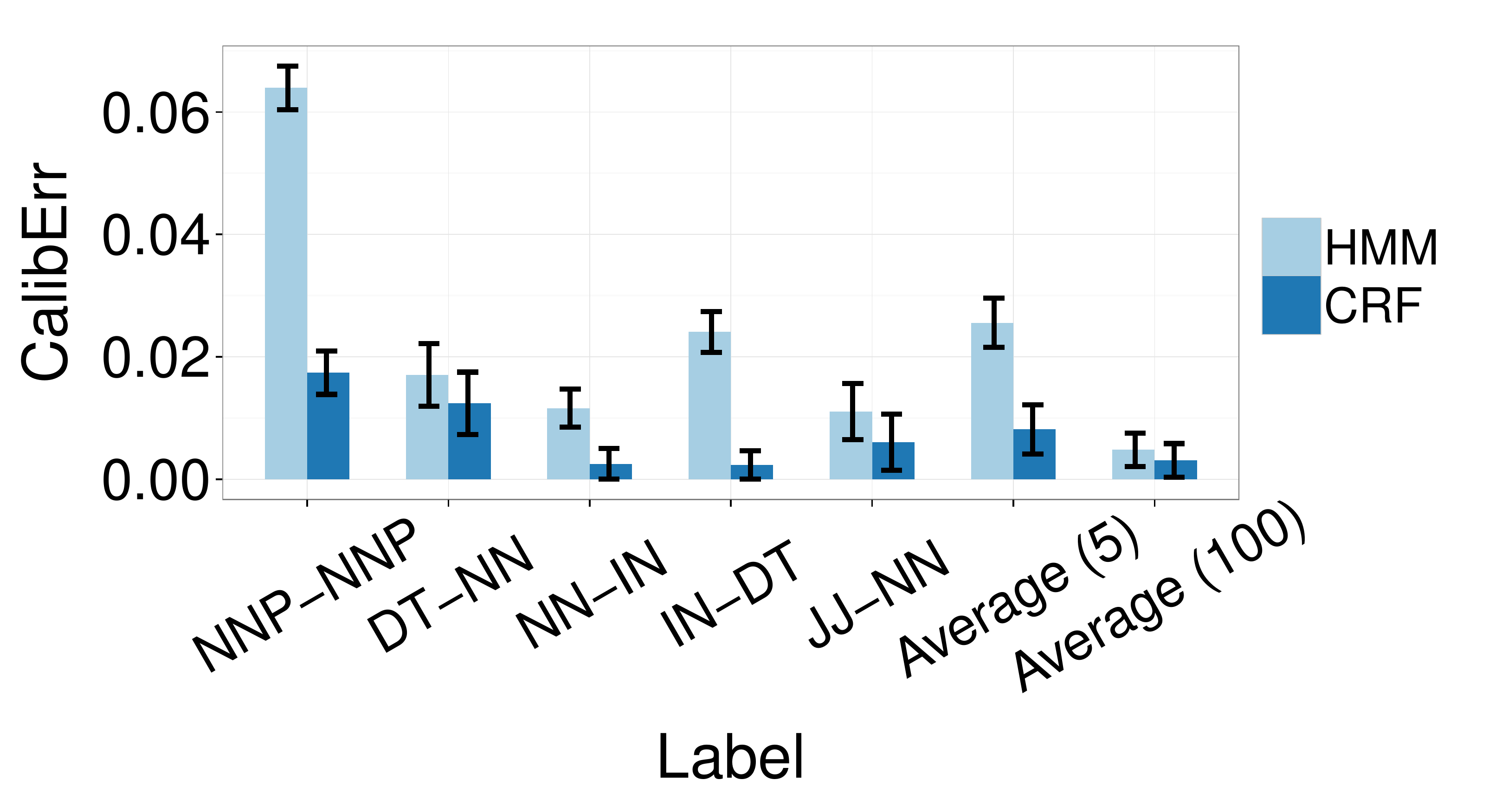}}
   \vspace{-0.1in}
   \caption{Calibration errors of HMM and CRF on predicting (a) single-word tags and (b) two-consecutive-word tags. Lower errors are better. The last two columns in each graph are the average calibration errors over the most common labels.  \vspace{-0.15in}}
   \label{fig:pos_score}
\end{figure}

There is no reason to restrict ourselves to model predictions of single words;
these models define marginal distributions over larger textual units.
Next we examine the calibration of posterior predictions of tag pairs on two consecutive words
in the test set. The same analysis may be important for, say, phrase extraction or other chunking/parsing tasks.

We report results for the top 5 and 100 most frequent tag pairs (Figure \ref{fig:pos_score}(b)). We observe a similar pattern as seen from the experiment on single tags:
the CRF is generally better calibrated than the HMM,
but the HMM does achieve better calibration errors in 29 out of 100 categories.

These tagging experiments illustrate that, depending on the application, different models can exhibit different levels of calibration.

\section{Coreference resolution}
\label{s:coref}
              
We examine a third model, a probabilistic model for within-document noun phrase coreference, which has an efficient sampling-based inference procedure.  
In this section we introduce it and analyze its calibration, in preparation for the next section where we use it for exploratory data analysis.

\subsection{Antecedent selection model}
We use the Berkeley coreference resolution system
\citep{durrett2013easy}, which was originally presented as a CRF;
we give it an equivalent
a series of independent logistic regressions (see appendix for details).
The primary component of this model
is a locally-normalized log-linear distribution
over clusterings of noun phrases, each cluster denoting an entity.  
The model takes a fixed input of $N$ mentions (noun phrases),
indexed by $i$ in their positional order in the document.
It posits that every mention $i$ has a latent antecedent selection decision, $a_i \in \{1,\dots,i-1, \textsc{new}\}$, denoting which previous mention it attaches to, or $\textsc{new}$ if it is starting a new entity that has not yet been seen at a previous position in the text.  Such a mention-mention attachment indicates coreference, while the final entity clustering includes more links implied through transitivity.
The model's generative process is:
\begin{defn}[Antencedent coreference model and sampling algorithm] \label{d:corefmodel}
\
\begin{itemizesquish}
\item For $i=1..N$, sample \\
$a_i \sim \frac{1}{Z_i} \exp( \mathbf{w}\transpose\mathbf{f}(i,a_i,x) )$
\item Calculate the entity clusters as $\mathbf{e} := CC(\mathbf{a})$, the connected components of the antecedent graph
having edges $(i, a_i)$ for $i$ where $a_i \neq \textsc{new}$.

\end{itemizesquish}
\end{defn}
\noindent
Here $x$ denotes all information in the document that is conditioned on for log-linear features $\mathbf{f}$.
$\mathbf{e}=\{e_1,...e_M\}$ denotes the entity clusters,
where each element is a set of mentions.
There are $M$ entity clusters corresponding to the number of connected components in $\mathbf{a}$.  
The model defines a joint distribution over antecedent decisions 
$P(\mathbf{a} | x) = \prod_i P(a_i | x)$;
it also defines a joint distribution
over entity clusterings 
$P(\mathbf{e}|x)$,
where the probability of an $\mathbf{e}$ is the sum of the probabilities of all
$\mathbf{a}$ vectors that could give rise to it.
In a manner similar to a
distance-dependent Chinese restaurant process \citep{Blei2011DDCRP},
it is non-parametric in the sense that the number of clusters $M$ is not fixed in advance.

\subsection{Sampling-based inference}
\label{s:corefsampling}

For both calibration analysis and exploratory applications, we need
to analyze the posterior distribution over entity clusterings.
This distribution is a complex mathematical object;
an attractive approach to analyze it is to
draw samples from this distribution, then analyze the samples.

This antecedent-based model
admits a very straightforward procedure to draw independent $\mathbf{e}$ samples,
by stepping through Def.~\ref{d:corefmodel}:
independently sample each $a_i$ then calculate the connected components of the 
resulting antecedent graph.
By construction, this procedure samples from the joint distribution of $\mathbf{e}$
(even though we never compute the probability of any single clustering $\mathbf{e}$).
 
Unlike approximate sampling approaches,
such as Markov chain Monte Carlo methods
used in other coreference work
to sample $\mathbf{e}$
\citep{Haghighi2007Coref},
here there are no questions about burn-in or autocorrelation \citep{Kass1998Roundtable}.
Every sample is independent and very fast to compute---only slightly slower than calculating the MAP assignment
(due to the $\exp$ and normalization for each $a_i$).  We implement this algorithm by modifying the publicly available implementation from \citeauthor{durrett2013easy}.\footnote{Berkeley Coreference Resolution System, version 1.1: \url{http://nlp.cs.berkeley.edu/projects/coref.shtml}}

\subsection{Calibration analysis}

\begin{figure}[t]
  \centering
  \vspace{-0.2in}
  \includegraphics[width=0.7\linewidth]{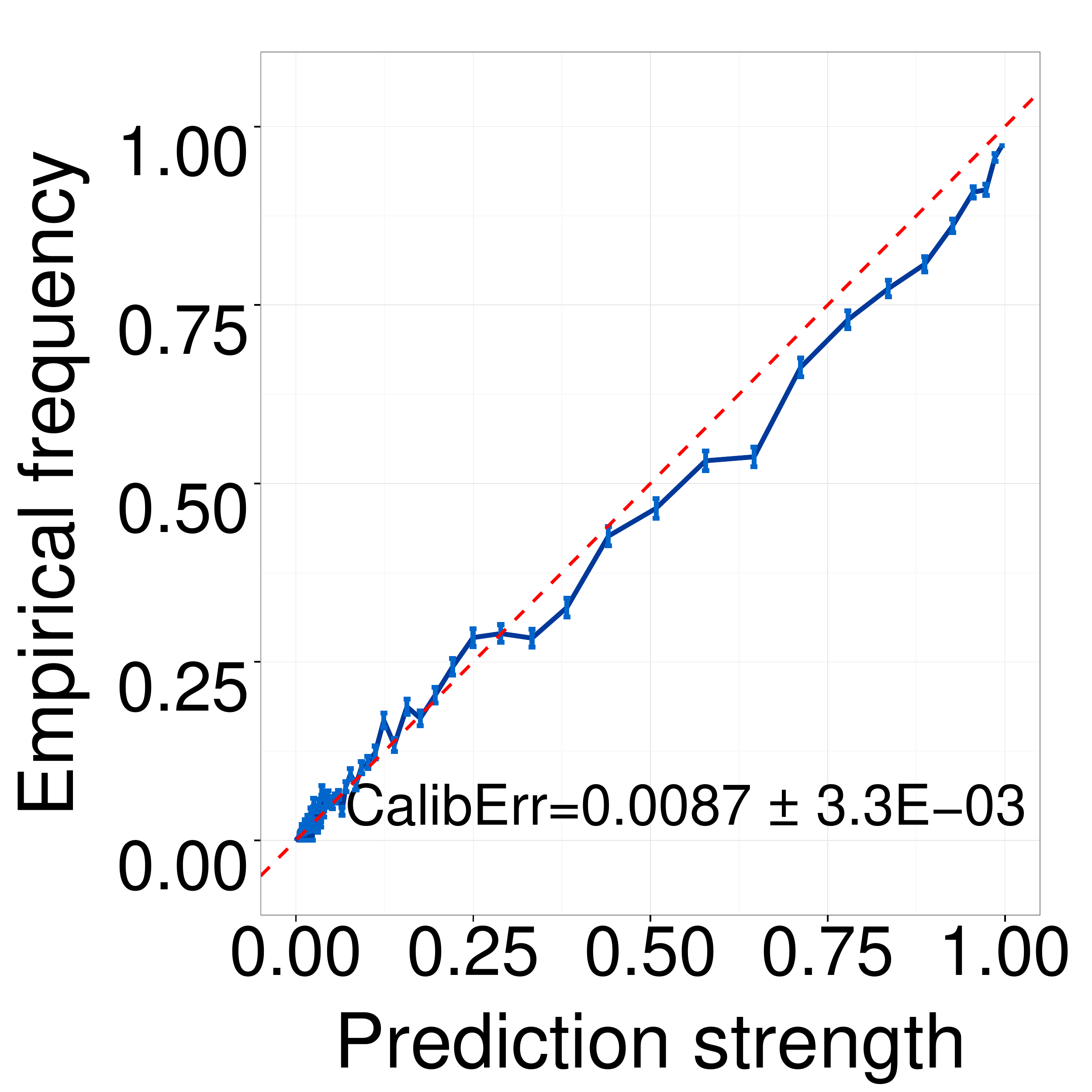}
  \caption{Coreference calibration curve
  for predicting whether two mentions belong to the same entity cluster. \vspace{-0.2in}}
  \label{fig:coref}
\end{figure}

We consider the following inference query:
for a randomly chosen pair of mentions,
are they coreferent?
Even if the model's accuracy is comparatively low,
it may be the case that it is correctly calibrated---if it thinks
there should be great variability in entity clusterings,
it may be uncertain whether a pair of mentions should belong together.

Let $\ell_{ij}$ be 1 if the mentions $i$ and $j$ are predicted to be coreferent, and 0 otherwise.
Annotated data defines a gold-standard $\ell^{(g)}_{ij}$ value for every pair $i,j$.
Any probability distribution over $\mathbf{e}$ defines a 
marginal Bernoulli distribution for every proposition $\ell_{ij}$,
marginalizing out $\mathbf{e}$:
\vspace{-0.05in}
\begin{equation}
  P( \ell_{ij}=1 \mid x) = 
\sum_{\mathbf{e}} 1\{ (i,j) \in \mathbf{e} \} \vspace{-0.05in}
P(\mathbf{e} \mid x) \end{equation}
where $(i,j) \in \mathbf{e}$ is true iff there is an entity in $\mathbf{e}$
that contains both $i$ and $j$.

In a traditional coreference evaluation of the best-prediction entity clustering,
the model assigns 1 or 0 to every $\ell_{ij}$ and the pairwise precision and recall can be computed by comparing them to the corresponding $\ell^{(g)}_{ij}$.
Here, we instead compare the $q_{ij} \equiv P(\ell_{ij}=1 \mid x, \mathbf{e})$ prediction strengths against $\ell^{(g)}_{ij}$ empirical frequencies
to assess pairwise calibration, 
with the same binary calibration analysis tools developed in \S\ref{s:empircalib}
by aggregating pairs with similar $q_{ij}$ values.
Each $q_{ij}$ is computed by averaging over
1,000 samples, simply taking the fraction of samples where the pair $(i, j)$ is coreferent.

We perform this analysis on the development section of the English CoNLL-2011
data (404 documents). Using the sampling inference method discussed in $\S\ref{s:corefsampling}$,
we compute 4.3 millions prediction-label pairs and measure their calibration error. Our result shows that the model produces very well-calibrated predictions with less than $1\%$ $CalibErr$ (Figure \ref{fig:coref}),
though slightly overconfident on middle to high-valued predictions.
The calibration error indicates that it is the most calibrated model we examine within this paper. This result suggests we might be able to trust its level of uncertainty.

                                                
                                 
\section{Uncertainty in Entity-based Exploratory Analysis}
\label{s:entevt}

\subsection{Entity-syntactic event aggregation}
We demonstrate one important use of calibration analysis:
to ensure the usefulness of
propagating uncertainty from coreference resolution
into a system for exploring unannotated text.
Accuracy cannot be calculated since there are no labels;
but if the system is calibrated,
we postulate that uncertainty information can help users
understand the underlying reliability of aggregated extractions
and isolate predictions that are more likely to contain errors.

We illustrate with an event analysis application
to count the number of ``country attack events'':
for a particular country of the world,
how many news articles describe an entity affiliated with that country
as the agent of an attack, and how does this number change over time?
This is a simplified version of a problem where such
systems have been built and used for political science analysis
\citep{Schrodt1994KEDS,Schrodt2012Review,Leetaru2013GDELT,Boschee2013,OConnor2013IR}.  
A coreference component can improve extraction coverage
in cases such as ``\textit{Russian troops} were sighted \dots and \textit{they attacked} \ldots''

We use the coreference system examined in \S\ref{s:coref} for this analysis.
To propagate coreference uncertainty, we re-run event extraction on multiple coreference samples generated from the algorithm described in \S\ref{s:corefsampling}, inducing a posterior distribution over the event counts.
To isolate the effects of coreference,
we use a very simple syntactic dependency system to identify affiliations and events.
Assume the availability of dependency parses for a document $d$, a coreference resolution $\mathbf{e}$, and a lexicon of country names, which contains a small set of words $w(c)$ for each country $c$; for example, $w(\text{FRA})= \{\text{france},\text{french}\}$.
The binary function $f(c,e; x_d)$ assesses
whether an entity $e$ is affiliated with country $c$ and is described as the agent of an attack,
based on document text and parses $x_d$;
$f$ returns true iff both:\footnote{Syntactic relations are Universal Dependencies \citep{DeMarneffe2014Universal}; more details for the extraction rules are in the appendix.}
\begin{itemizesquish}\vspace{-0mm}
\item There exists a mention $i \in e$ described as country $c$: either its head word is in $w(c)$ (e.g. ``Americans''), or its head word has an \emph{nmod} or \emph{amod} modifier in $w(c)$ (e.g. ``American forces'', ``president of the U.S.''); and there is only one unique country $c$ among the mentions in the entity.
\item There exists a mention $j \in e$ 
which is the
\emph{nsubj} or \emph{agent} argument
to the verb ``attack'' (e.g.~``they attacked'', ``the forces attacked'', 
``attacked by them'').
\end{itemizesquish}\vspace{-0mm}
\noindent
For a given $c$, we first calculate
a binary variable for
whether there is at least one entity fulfilling $f$ in a particular document,
\vspace{-0.1in} 
\begin{equation}
a(d, c, \mathbf{e}_d) = \bigvee_{e \in \mathbf{e}_d} f(c,e; x_d)
\vspace{-0.1in}
\end{equation}
and second, the number of such documents
in $d(t)$, the set of \emph{New York Times} articles
published in a given time period $t$,
\vspace{-0.1in} 
\begin{equation}
n(t, c, \mathbf{e}_{d(t)}) = \sum_{d \in d(t)} a(d, c, \mathbf{e}_d) 
\vspace{-0.1in}
\end{equation}
These quantities are both random variables, since they depend on $\mathbf{e}$;
thus we are interested in the posterior distribution of $n$, marginalizing out $\mathbf{e}$,
\begin{align} \label{e:ntc} 
P(n(t, c, \mathbf{e}_{d(t)}) \mid x_{d(t)})
\end{align}

\noindent If our coreference model was highly certain (only one structure, or a small number of similar structures, had most of the probability mass in the space of all possible structures),
each document would have an $a$ posterior near either 0 or 1, and their sum in Eq.~\ref{e:ntc} would have
a narrow distribution.  
But if the model is uncertain, the distribution will be wider.
Because of the transitive closure,
the probability of $a$ is potentially more complex than the single antecedent linking probability between two mentions---the affiliation and attack information can propagate
through a long coreference chain.

\subsection{Results}

We tag and parse
a
193,403 article subset
of the Annotated New York Times LDC corpus
\citep{SandhausNYT}, which includes articles
about world news
from the years 1987 to 2007 (details in appendix).
For each article, we run the coreference system
to predict 100 samples, and evaluate $f$
on every entity in every sample.\footnote{We obtained similar results using only 10 samples. We also obtained similar results with a different query function, the total number of entities, across documents, that fulfill $f$.}
The quantity of interest is the number of articles mentioning attacks in a 3-month period (quarter),
for a given country.
Figure~\ref{f:timeseries} illustrates the 
mean and 95\% posterior credible intervals for each quarter.
The posterior mean $m$ is calculated as the mean of the samples,
and the interval is the normal approximation $m \pm 1.96\ s$, where $s$ is the standard deviation among samples for that country and time period.


Uncertainty information helps us understand whether a difference between data points is real.
In the plots of Figure~\ref{f:timeseries}, if we had used a 1-best coreference resolution, only a single line would be shown, with no assessment of uncertainty.
This is problematic in cases when the model genuinely does not know the correct answer.
For example, the 1993-1996 period of the USA plot (Figure~\ref{f:timeseries}, top) shows the posterior mean fluctuating from 1 to 5 documents; but when credible intervals are taken into consideration, we see that model does not know whether the differences are real, or were caused by coreference noise.

A similar case is highlighted at the bottom plot of Figure~\ref{f:timeseries}. 
Here we compare the event counts for Yugoslavia and NATO,
which were engaged in a conflict in 1999.  Did the \emph{New York Times} devote more attention to the attacks by one particular side? To a 1-best system, the answer would be yes. But the posterior intervals for the two countries' event counts in mid-1999 heavily overlap, indicating that the coreference system introduces too much uncertainty to obtain a conclusive answer for this question. 
Note that calibration of the coreference model is important for the credible intervals to be useful; for example, if the model was badly calibrated by being overconfident (too much probability over a small set of similar structures), these intervals would be too narrow, leading to incorrect interpretations of the event dynamics.

Visualizing this uncertainty gives richer information for a potential user of an NLP-based system, compared to simply drawing a line based on a single 1-best prediction.  It preserves the genuine uncertainty due to ambiguities the system was unable to resolve.  This highlights an alternative
use of \cite{Finkel2006Pipeline}'s approach of sampling multiple NLP pipeline components, which in that work was used to perform joint inference.  Instead of focusing on improving an NLP pipeline, we can pass uncertainty on to exploratory purposes, and try to highlight to a user where the NLP system may be wrong, or where it can only imprecisely specify a quantity of interest.

Finally, calibration can help error analysis.
For a calibrated model, the more uncertain a prediction is, the more likely it is to be erroneous. 
While coreference errors comprise only one part of event extraction errors
(alongside issues in parse quality, factivity, semantic roles, etc.),
we can look at highly uncertain event predictions
to understand the nature of coreference errors relative to our task.
We manually analyzed documents with a 50\% probability to contain an ``attack''ing country-affiliated entity, and found difficult coreference cases. 

In one article from late 1990,
an ``attack'' event for IRQ is extracted from the sentence
``But some political leaders said that they feared that \textit{Mr.~Hussein might attack} Saudi Arabia''.
The mention ``Mr.\ Hussein'' is classified as IRQ only when
it is coreferent with a previous mention
``President Saddam Hussein of Iraq'';
this occurs only 50\% of the time, since in some posterior samples
the coreference system split apart these two ``Hussein'' mentions.
This particular document is additionally difficult, since it includes the names of more than 10 countries (e.g.~United States, Saudi Arabia, Egypt), and some of the Hussein mentions are even clustered with presidents of other countries (such as ``President Bush''),
presumably because they share the ``president'' title.
These types of errors are a major issue for 
a political analysis task;
further analysis could assess their prevalence and how to address them in future work.


\newcommand{\tspdf}[1]{\includegraphics[width=3.0in]{#1}}

\begin{figure}[t]
\vspace{0in}
\tspdf{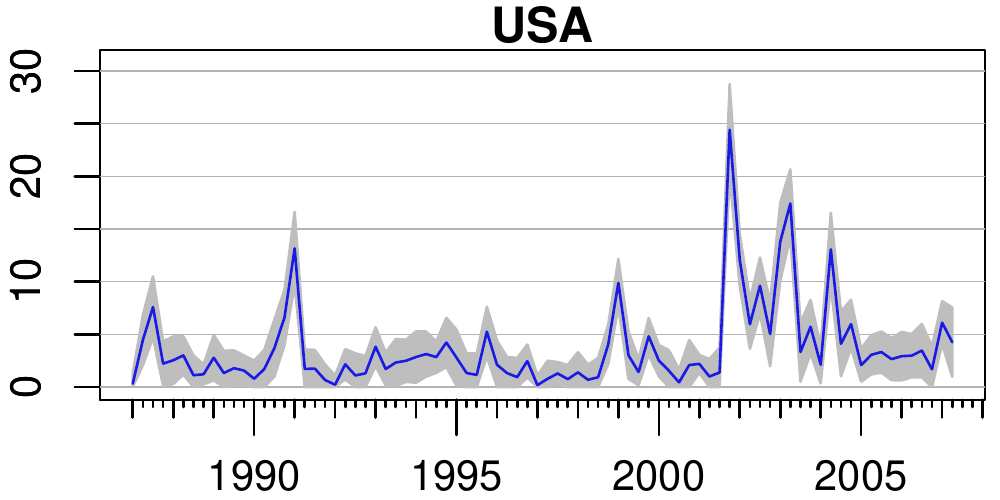}
\tspdf{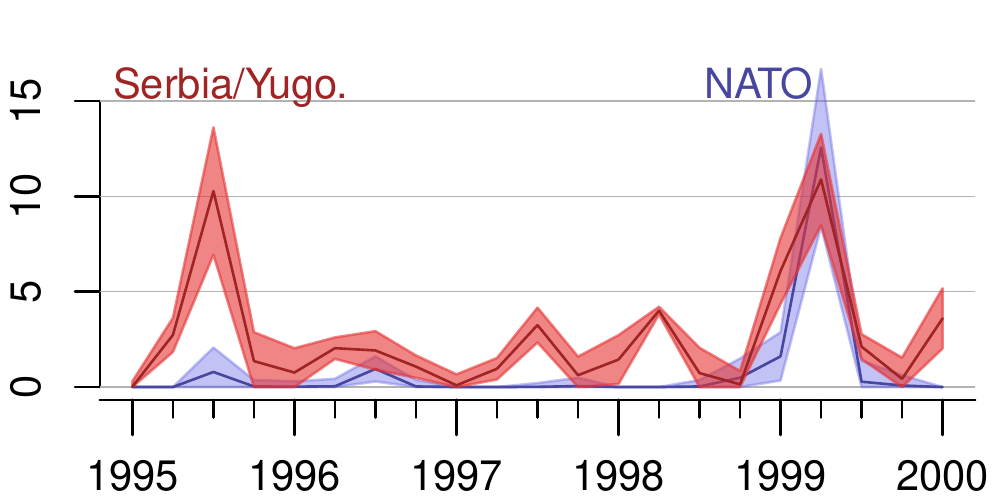}
\caption{Number of documents with an ``attack''ing country per 3-month period,
and coreference posterior uncertainty for that quantity.
The dark line is the posterior mean, and the shaded region is the 95\% posterior credible interval.
See appendix for more examples.
\label{f:timeseries}}
\vspace{-0mm}
\end{figure}

\section{Conclusion}

In this work, we argue that the calibration of posterior predictions is a desirable property of probabilistic NLP models,
and that it can be directly evaluated.
We also demonstrate a use case of having calibrated uncertainty: its propagation into downstream exploratory analysis.

Our posterior simulation approach for exploratory and error analysis
relates to \emph{posterior predictive checking}
\citep{Gelman2013BDA},
which analyzes a posterior to test model assumptions;
\cite{Mimno2011PPC} apply it to a topic model.

One avenue of future work is to investigate more effective nonparametric regression methods
to better estimate and visualize calibration error, such as Gaussian processes or bootstrapped kernel density estimation.

Another important question is: what types of inferences are facilitated by correct calibration?
Intuitively, we think that overconfidence will lead to overly narrow confidence intervals;
but in what sense are confidence intervals ``good'' when calibration is perfect?
Also, does calibration help joint inference in NLP pipelines?
It may also assist calculations that rely on expectations,
such as inference methods like minimum Bayes risk decoding,
or learning methods like EM,
since calibrated predictions imply that calculated expectations are statistically unbiased
(though the implications of this fact may be subtle).
Finally, it may be interesting
to pursue recalibration methods, which readjust a non-calibrated model's predictions
to be calibrated; recalibration methods have been developed
for 
binary \citep{Platt1999Isotonic,Niculescu2005Calibration}
and multiclass \citep{Zadrozny2002MulticlassCalib}
classification settings,
but we are unaware of methods appropriate
for the highly structured outputs typical in linguistic analysis.
Another approach might be to directly constrain $CalibErr=0$ during training,
or try to reduce it as a training-time risk minimization or cost objective
\citep{Smith2006Risk,Gimpel2010Softmax,Stoyanov2011Risk,Brummer2013Scoring}.

Calibration is an interesting and important property of NLP models.
Further work is necessary to address these and many other questions.

\section*{Acknowledgments}
Thanks to Erik Learned-Miller, Benjamin Marlin, Craig Greenberg, Phan-Minh Nguyen, Caitlin Cellier and the CMU ARK Lab for discussion and comments,
and to the anonymous reviewers (especially R3) for helpful suggestions.

\bibliographystyle{plainnat}
\bibliography{emnlp2015,brenocon}

\begin{thebibliography}{47}
\providecommand{\natexlab}[1]{#1}
\providecommand{\url}[1]{\texttt{#1}}
\expandafter\ifx\csname urlstyle\endcsname\relax
  \providecommand{\doi}[1]{doi: #1}\else
  \providecommand{\doi}{doi: \begingroup \urlstyle{rm}\Url}\fi

\bibitem[Bamman et~al.(2013)Bamman, {O'Connor}, and Smith]{Bamman2013Personas}
David Bamman, Brendan {O'Connor}, and Noah~A. Smith.
\newblock Learning latent personas of film characters.
\newblock In \emph{Proceedings of ACL}, 2013.

\bibitem[Bennett(2000)]{Bennett2000NB}
Paul~N. Bennett.
\newblock Assessing the calibration of naive {B}ayes' posterior estimates.
\newblock Technical report, Carnegie Mellon University, 2000.

\bibitem[Blei and Frazier(2011)]{Blei2011DDCRP}
David~M. Blei and Peter~I. Frazier.
\newblock Distance dependent {C}hinese restaurant processes.
\newblock \emph{The Journal of Machine Learning Research}, 12:\penalty0
  2461--2488, 2011.

\bibitem[Boschee et~al.(2013)Boschee, Natarajan, and Weischedel]{Boschee2013}
Elizabeth Boschee, Premkumar Natarajan, and Ralph Weischedel.
\newblock Automatic extraction of events from open source text for predictive
  forecasting.
\newblock \emph{Handbook of Computational Approaches to Counterterrorism},
  page~51, 2013.

\bibitem[Brier(1950)]{Brier1950}
Glenn~W. Brier.
\newblock Verification of forecasts expressed in terms of probability.
\newblock \emph{Monthly weather review}, 78\penalty0 (1):\penalty0 1--3, 1950.

\bibitem[Br{\"o}cker(2009)]{Brocker2009Decomposition}
Jochen Br{\"o}cker.
\newblock Reliability, sufficiency, and the decomposition of proper scores.
\newblock \emph{Quarterly Journal of the Royal Meteorological Society},
  135\penalty0 (643):\penalty0 1512--1519, 2009.

\bibitem[Br{\"u}mmer and Doddington(2013)]{Brummer2013Scoring}
Niko Br{\"u}mmer and George Doddington.
\newblock Likelihood-ratio calibration using prior-weighted proper scoring
  rules.
\newblock \emph{arXiv preprint arXiv:1307.7981}, 2013.
\newblock Interspeech 2013.

\bibitem[de~Marneffe et~al.(2014)de~Marneffe, Dozat, Silveira, Haverinen,
  Ginter, Nivre, and Manning]{DeMarneffe2014Universal}
Marie-Catherine de~Marneffe, Timothy Dozat, Natalia Silveira, Katri Haverinen,
  Filip Ginter, Joakim Nivre, and Christopher~D. Manning.
\newblock Universal {S}tanford dependencies: A cross-linguistic typology.
\newblock In \emph{Proceedings of LREC}, 2014.

\bibitem[DeGroot and Fienberg(1983)]{degroot1983comparison}
Morris~H. DeGroot and Stephen~E. Fienberg.
\newblock The comparison and evaluation of forecasters.
\newblock \emph{The statistician}, pages 12--22, 1983.

\bibitem[Domingos and Pazzani(1997)]{Domingos1997NB}
Pedro Domingos and Michael Pazzani.
\newblock On the optimality of the simple {B}ayesian classifier under zero-one
  loss.
\newblock \emph{Machine learning}, 29\penalty0 (2-3):\penalty0 103--130, 1997.

\bibitem[Durrett and Klein(2013)]{durrett2013easy}
Greg Durrett and Dan Klein.
\newblock Easy victories and uphill battles in coreference resolution.
\newblock In \emph{EMNLP}, pages 1971--1982, 2013.

\bibitem[Durrett and Klein(2014)]{durrett2014joint}
Greg Durrett and Dan Klein.
\newblock A joint model for entity analysis: Coreference, typing, and linking.
\newblock \emph{Transactions of the Association for Computational Linguistics},
  2:\penalty0 477--490, 2014.

\bibitem[Finkel et~al.(2006)Finkel, Manning, and Ng]{Finkel2006Pipeline}
Jenny~Rose Finkel, Christopher~D. Manning, and Andrew~Y. Ng.
\newblock Solving the problem of cascading errors: Approximate {B}ayesian
  inference for linguistic annotation pipelines.
\newblock In \emph{Proceedings of the 2006 Conference on Empirical Methods in
  Natural Language Processing}, pages 618--626. Association for Computational
  Linguistics, 2006.

\bibitem[Gelman et~al.(2013)Gelman, Carlin, Stern, Dunson, Vehtari, and
  Rubin]{Gelman2013BDA}
Andrew Gelman, John~B. Carlin, Hal~S. Stern, David~B. Dunson, Aki Vehtari, and
  Donald~B. Rubin.
\newblock \emph{Bayesian data analysis}.
\newblock Chapman and Hall/CRC, 3rd edition, 2013.

\bibitem[Gimpel and Smith(2008)]{Gimpel2008MT}
Kevin Gimpel and Noah~A. Smith.
\newblock Rich source-side context for statistical machine translation.
\newblock In \emph{Proceedings of the Third Workshop on Statistical Machine
  Translation}, pages 9--17, 2008.

\bibitem[Gimpel and Smith(2010)]{Gimpel2010Softmax}
Kevin Gimpel and Noah~A. Smith.
\newblock Softmax-margin {CRF}s: Training log-linear models with cost
  functions.
\newblock In \emph{Human Language Technologies: The 2010 Annual Conference of
  the North American Chapter of the Association for Computational Linguistics},
  pages 733--736. Association for Computational Linguistics, 2010.

\bibitem[Gimpel et~al.(2013)Gimpel, Batra, Dyer, and
  Shakhnarovich]{gimpel2013diversity}
Kevin Gimpel, Dhruv Batra, Chris Dyer, and Gregory Shakhnarovich.
\newblock A systematic exploration of diversity in machine translation.
\newblock In \emph{Proceedings of the 2013 Conference on Empirical Methods in
  Natural Language Processing}, pages 1100--1111, Seattle, Washington, USA,
  October 2013. Association for Computational Linguistics.
\newblock URL \url{http://www.aclweb.org/anthology/D13-1111}.

\bibitem[Gneiting and Raftery(2007)]{Gneiting2007Scoring}
Tilmann Gneiting and Adrian~E. Raftery.
\newblock Strictly proper scoring rules, prediction, and estimation.
\newblock \emph{Journal of the American Statistical Association}, 102\penalty0
  (477):\penalty0 359--378, 2007.

\bibitem[Goodman(1996)]{Goodman1996MBR}
Joshua Goodman.
\newblock Parsing algorithms and metrics.
\newblock In \emph{Proceedings of the 34th Annual Meeting of the Association
  for Computational Linguistics}, pages 177--183, Santa Cruz, California, USA,
  June 1996. Association for Computational Linguistics.
\newblock \doi{10.3115/981863.981887}.
\newblock URL \url{http://www.aclweb.org/anthology/P96-1024}.

\bibitem[Haghighi and Klein(2007)]{Haghighi2007Coref}
Aria Haghighi and Dan Klein.
\newblock Unsupervised coreference resolution in a nonparametric {B}ayesian
  model.
\newblock In \emph{Annual Meeting, Association for Computational Linguistics},
  volume~45, page 848, 2007.

\bibitem[Kass et~al.(1998)Kass, Carlin, Gelman, and Neal]{Kass1998Roundtable}
Robert~E. Kass, Bradley~P. Carlin, Andrew Gelman, and Radford~M. Neal.
\newblock Markov chain {M}onte {C}arlo in practice: a roundtable discussion.
\newblock \emph{The American Statistician}, 52\penalty0 (2):\penalty0 93--100,
  1998.

\bibitem[Kumar and Byrne(2004)]{Kumar200MBR}
Shankar Kumar and William Byrne.
\newblock Minimum {B}ayes-risk decoding for statistical machine translation.
\newblock In Daniel~Marcu Susan~Dumais and Salim Roukos, editors,
  \emph{HLT-NAACL 2004: Main Proceedings}, pages 169--176, Boston,
  Massachusetts, USA, May 2 - May 7 2004. Association for Computational
  Linguistics.

\bibitem[Leetaru and Schrodt(2013)]{Leetaru2013GDELT}
Kalev Leetaru and Philip~A. Schrodt.
\newblock {GDELT}: Global data on events, location, and tone, 1979--2012.
\newblock In \emph{ISA Annual Convention}, volume~2, page~4, 2013.

\bibitem[Lin and Kolcz(2012)]{Lin2012Twitter}
Jimmy Lin and Alek Kolcz.
\newblock Large-scale machine learning at {T}witter.
\newblock In \emph{Proceedings of the 2012 ACM SIGMOD International Conference
  on Management of Data}, pages 793--804. ACM, 2012.

\bibitem[McCord et~al.(2012)McCord, Murdock, and Boguraev]{McCord2012Parsing}
Michael~C. McCord, J.~William Murdock, and Branimir~K. Boguraev.
\newblock Deep parsing in {W}atson.
\newblock \emph{IBM Journal of Research and Development}, 56\penalty0
  (3.4):\penalty0 3--1, 2012.

\bibitem[Mimno and Blei(2011)]{Mimno2011PPC}
David Mimno and David Blei.
\newblock Bayesian checking for topic models.
\newblock In \emph{Proceedings of the 2011 Conference on Empirical Methods in
  Natural Language Processing}, pages 227--237, Edinburgh, Scotland, UK., July
  2011. Association for Computational Linguistics.
\newblock URL \url{http://www.aclweb.org/anthology/D11-1021}.

\bibitem[Miwa et~al.(2010)Miwa, Pyysalo, Hara, and
  Tsujii]{Miwa2010Dependencies}
Makoto Miwa, Sampo Pyysalo, Tadayoshi Hara, and Jun'ichi Tsujii.
\newblock Evaluating dependency representations for event extraction.
\newblock In \emph{Proceedings of the 23rd International Conference on
  Computational Linguistics (Coling 2010)}, pages 779--787, Beijing, China,
  August 2010. Coling 2010 Organizing Committee.
\newblock URL \url{http://www.aclweb.org/anthology/C10-1088}.

\bibitem[Murphy and Winkler(1987)]{Murphy1987General}
Allan~H. Murphy and Robert~L. Winkler.
\newblock A general framework for forecast verification.
\newblock \emph{Monthly Weather Review}, 115\penalty0 (7):\penalty0 1330--1338,
  1987.

\bibitem[Ng and Jordan(2002)]{Ng2002NBLR}
Andrew Ng and Michael Jordan.
\newblock On discriminative vs. generative classifiers: A comparison of
  logistic regression and naive {B}ayes.
\newblock \emph{Advances in neural information processing systems},
  14:\penalty0 841, 2002.

\bibitem[Niculescu-Mizil and Caruana(2005)]{Niculescu2005Calibration}
Alexandru Niculescu-Mizil and Rich Caruana.
\newblock Predicting good probabilities with supervised learning.
\newblock In \emph{Proceedings of the 22nd {I}nternational {C}onference on
  {M}achine {L}earning}, pages 625--632, 2005.

\bibitem[{O'Connor} et~al.(2013){O'Connor}, Stewart, and Smith]{OConnor2013IR}
Brendan {O'Connor}, Brandon Stewart, and Noah~A. Smith.
\newblock Learning to extract international relations from political context.
\newblock In \emph{Proceedings of ACL}, 2013.

\bibitem[Okazaki(2007)]{CRFsuite}
Naoaki Okazaki.
\newblock Crfsuite: a fast implementation of conditional random fields
  ({CRF}s), 2007.
\newblock URL \url{http://www.chokkan.org/software/crfsuite/}.

\bibitem[Pedregosa et~al.(2011)Pedregosa, Varoquaux, Gramfort, Michel, Thirion,
  Grisel, Blondel, Prettenhofer, Weiss, Dubourg, Vanderplas, Passos,
  Cournapeau, Brucher, Perrot, and Duchesnay]{scikit-learn}
F.~Pedregosa, G.~Varoquaux, A.~Gramfort, V.~Michel, B.~Thirion, O.~Grisel,
  M.~Blondel, P.~Prettenhofer, R.~Weiss, V.~Dubourg, J.~Vanderplas, A.~Passos,
  D.~Cournapeau, M.~Brucher, M.~Perrot, and E.~Duchesnay.
\newblock Scikit-learn: Machine learning in {P}ython.
\newblock \emph{Journal of Machine Learning Research}, 12:\penalty0 2825--2830,
  2011.

\bibitem[Platt(1999)]{Platt1999Isotonic}
John Platt.
\newblock Probabilistic outputs for support vector machines and comparisons to
  regularized likelihood methods.
\newblock In \emph{Advances in large margin classifiers}. MIT Press (2000),
  1999.
\newblock URL \url{http://research.microsoft.com/pubs/69187/svmprob.ps.gz}.

\bibitem[Pradhan et~al.(2011)Pradhan, Ramshaw, Marcus, Palmer, Weischedel, and
  Xue]{pradhan2011conll}
Sameer Pradhan, Lance Ramshaw, Mitchell Marcus, Martha Palmer, Ralph
  Weischedel, and Nianwen Xue.
\newblock {CoNLL}-2011 shared task: Modeling unrestricted coreference in
  {O}ntonotes.
\newblock In \emph{Proceedings of the Fifteenth Conference on Computational
  Natural Language Learning: Shared Task}, pages 1--27. Association for
  Computational Linguistics, 2011.

\bibitem[Read(2005)]{Read2005Emoticons}
Jonathon Read.
\newblock Using emoticons to reduce dependency in machine learning techniques
  for sentiment classification.
\newblock In \emph{Proceedings of the ACL Student Research Workshop}, pages
  43--48. Association for Computational Linguistics, 2005.

\bibitem[Sandhaus(2008)]{SandhausNYT}
Evan Sandhaus.
\newblock The {Ne}w {Y}ork {T}imes {A}nnotated {C}orpus.
\newblock \emph{Linguistic Data Consortium}, LDC2008T19, 2008.

\bibitem[Schrodt(2012)]{Schrodt2012Review}
Philip~A. Schrodt.
\newblock Precedents, progress, and prospects in political event data.
\newblock \emph{International Interactions}, 38\penalty0 (4):\penalty0
  546--569, 2012.

\bibitem[Schrodt et~al.(1994)Schrodt, Davis, and Weddle]{Schrodt1994KEDS}
Philip~A. Schrodt, Shannon~G. Davis, and Judith~L. Weddle.
\newblock {KEDS} -- a program for the machine coding of event data.
\newblock \emph{Social Science Computer Review}, 12\penalty0 (4):\penalty0 561
  --587, December 1994.
\newblock \doi{10.1177/089443939401200408}.
\newblock URL \url{http://ssc.sagepub.com/content/12/4/561.abstract}.

\bibitem[Singh et~al.(2013)Singh, Riedel, Martin, Zheng, and
  McCallum]{singh2013joint}
Sameer Singh, Sebastian Riedel, Brian Martin, Jiaping Zheng, and Andrew
  McCallum.
\newblock Joint inference of entities, relations, and coreference.
\newblock In \emph{Proceedings of the 2013 Workshop on Automated Knowledge Base
  Construction}, pages 1--6. ACM, 2013.

\bibitem[Smith and Eisner(2006)]{Smith2006Risk}
David~A. Smith and Jason Eisner.
\newblock Minimum risk annealing for training log-linear models.
\newblock In \emph{Proceedings of the COLING/ACL 2006 Main Conference Poster
  Sessions}, pages 787--794, Sydney, Australia, July 2006. Association for
  Computational Linguistics.
\newblock URL \url{http://www.aclweb.org/anthology/P06-2101}.

\bibitem[Stoyanov et~al.(2011)Stoyanov, Ropson, and Eisner]{Stoyanov2011Risk}
Veselin Stoyanov, Alexander Ropson, and Jason Eisner.
\newblock Empirical risk minimization of graphical model parameters given
  approximate inference, decoding, and model structure.
\newblock In \emph{International Conference on Artificial Intelligence and
  Statistics}, pages 725--733, 2011.

\bibitem[Toutanova et~al.(2008)Toutanova, Haghighi, and
  Manning]{toutanova2008global}
Kristina Toutanova, Aria Haghighi, and Christopher~D. Manning.
\newblock A global joint model for semantic role labeling.
\newblock \emph{Computational Linguistics}, 34\penalty0 (2):\penalty0 161--191,
  2008.

\bibitem[Tukey(1961)]{Tukey1961Regressogram}
John~W. Tukey.
\newblock Curves as parameters, and touch estimation.
\newblock In \emph{Proceedings of the Fourth Berkeley Symposium on Mathematical
  Statistics and Probability, Volume 1: Contributions to the Theory of
  Statistics}, pages 681--694, Berkeley, Calif., 1961. University of California
  Press.
\newblock URL \url{http://projecteuclid.org/euclid.bsmsp/1200512189}.

\bibitem[Venugopal et~al.(2008)Venugopal, Zollmann, Smith, and
  Vogel]{venugopal2008pipelines}
Ashish Venugopal, Andreas Zollmann, Noah~A. Smith, and Stephan Vogel.
\newblock Wider pipelines: N-best alignments and parses in {MT} training.
\newblock In \emph{Proceedings of AMTA}, 2008.

\bibitem[Wasserman(2006)]{Wasserman2006AllNP}
Larry Wasserman.
\newblock \emph{All of nonparametric statistics}.
\newblock Springer Science \& Business Media, 2006.

\bibitem[Zadrozny and Elkan(2002)]{Zadrozny2002MulticlassCalib}
Bianca Zadrozny and Charles Elkan.
\newblock Transforming classifier scores into accurate multiclass probability
  estimates.
\newblock In \emph{Proceedings of {KDD}}, pages 694--699. ACM, 2002.

\end{thebibliography}

\newpage
\renewcommand{\partname}{}
\renewcommand{\thepart}{}
\part{\text{\Large Appendix}}

\title{
    Supplementary information for
    ``Posterior calibration and exploratory analysis
    for natural language processing models'' (EMNLP 2015)
}

\maketitle

\setcounter{section}{0}
\section{Sampling a deterministic function of a random variable}
In several places in this paper, we define probability distributions
over deterministic functions of a random variable, and sample from them
by applying the deterministic function to samples of the random variable.
This should be valid by construction, but we supply the following argument
for further justification.

$X$ is a random variable and $g(x)$ is a deterministic function which takes a value of $X$ as its input.
Since $g$ depends on a random variable, $g(X)$ is a random variable as well.
The distribution for $g(X)$, or aspects of it (such as a PMF or independent samples from it) can be calculated by marginalizing out $X$ with a Monte Carlo approximation.  Assuming $g$ has discrete outputs (as is the case for the event counting function $n$, or connected components function $CC$), we examine the probability mass function:
\begin{align}
\text{pmf}(h) &\equiv P(g(X)=h) \\
&= \sum_x P(g(x) = h \mid x)\ P(x)  \\
& = \sum_x 1\{g(x)=h\} P(x)  \label{e:determ} \\
& \approx \frac{1}{S} \sum_{x \sim P(X)} 1\{g(x)=h\} \label{e:determsamp}
\end{align}
Eq.~\ref{e:determ} holds because $g(x)$ is a deterministic function,
and Eq.~\ref{e:determsamp} is a Monte Carlo approximation that uses $S$ samples from $P(x)$.

This implies that a set of $g$ values calculated on $x$ samples,
$\{g(x^{(s)}) : x^{(s)} \sim P(x)\}$, should constitute a sample from the distribution $P(g(X))$; in our event analysis section we usually call this the ``posterior'' distribution of $g(X)$ (the $n(t,c)$ function there).
In our setting, we do not directly use the PMF calculation above;
instead, we construct normal approximations to the probability distribution $g(X)$.

We use this technique in several places.
For the calibration error confidence interval, the calibration error is 
a deterministic function of the uncertain empirical label frequencies $p_i$;
there, we propagate posterior uncertainty from a normal approximation to the Bernoulli parameter's posterior (the $p_i$ distribution under the central limit theorem)
through simulation.
In the coreference model, the connected components function is a deterministic function of the antecedent vector;
thus repeatedly calculating $\mathbf{e}^{(s)} := CC(\mathbf{a}^{(s)})$ yields samples of entity clusterings from their posterior.
For the event analysis, the counting function $n(t, c, \mathbf{e}_{d(t)})$ is a function of the entity samples, and thus can be recalculated on each---this is a multiple step deterministic pipeline, which postprocesses simulated random variables.

As in other Monte Carlo-based inference techniques (as applied to both Bayesian and frequentist (e.g.~bootstrapping) inference), the mean and standard deviation of samples drawn from the distribution constitute the mean and standard deviation of the desired posterior distribution,
subject to Monte Carlo error due to the finite number of samples,
which by the central limit theorem shrinks at a rate of $1/\sqrt{S}$.
The Monte Carlo standard error for estimating the mean is $\sigma/\sqrt{S}$ where $\sigma$ is the standard deviation.
So with 100 samples, the Monte Carlo standard error for the mean is $\sqrt{100}=10$ times smaller than standard deviation.  Thus in the time series graphs, which are based on $S=100$ samples, the posterior mean (dark line) has Monte Carlo uncertainty that is 10 times smaller than the vertical gray area (95\% CI) around it.

\section{Normalization in the coreference model}

\noindent
Durrett and Klein (2013) present their model as a globally normalized, but fully factorized, CRF:
\[ P(\mathbf{a}|x) = \frac{1}{Z} \prod_i \exp(\mathbf{w}\transpose\mathbf{f}(i,a_i,x)) \]
Since the factor function decomposes independently for each random variable $a_i$,
their probabilities are actually independent, and can be rewritten
with local normalization,
\[ P(\mathbf{a}|x) = \prod_i \frac{1}{Z_i} \exp(\mathbf{w}\transpose\mathbf{f}(i,a_i,x)) \]

\noindent
This interpretation justifies the use of independent sampling
to draw samples of the joint posterior.

\section{Event analysis: Corpus selection, country affiliation, and parsing}

Articles are filtered to yield a dataset about world news.  In the New York Times Annotated Corpus, every article is tagged with a large set of labels.  We include
articles that contain a category whose label starts with the string \emph{Top/News/World},
and exclude articles with any category matching the regex \emph{/(Sports$|$Opinion)},
and whose text body contains a mention of at least one country name.

Country names are taken from the dictionary \emph{country\_igos.txt} 
based on previous work
(\url{http://brenocon.com/irevents/}).  Country name matching is case insensitive and uses light stemming: when trying to match a word against the lexicon, if a match is not found, it backs off to stripping the last and last two characters.  (This is usually unnecessary since the dictionary contains modifier forms.)

POS, NER, and constituent and dependency parses are produced with Stanford CoreNLP 3.5.2 with default settings except for one change, to use its shift-reduce constituent parser (for convenience of processing speed).
We treat tags and parses as fixed and leave their uncertainty propagation for future work.

When formulating the extraction rules, we examined frequencies of all syntactic dependencies within country-affiliated entities, in order to help find reasonably high-coverage syntactic relations for the ``attack'' rule.

\section{Event time series graphs}
The following pages contain posterior time series graphs
for 20 countries, as described in the section on coreference-based
event aggregation, in order of decreasing total event frequency.
As in the main paper, the blue line indicates the posterior mean,
and the gray region indicates 95\% posterior credibility intervals,
with count aggregation at the monthly level.
The titles are ISO3 country codes.

\newpage
\tspdf{1_USA.pdf}
\tspdf{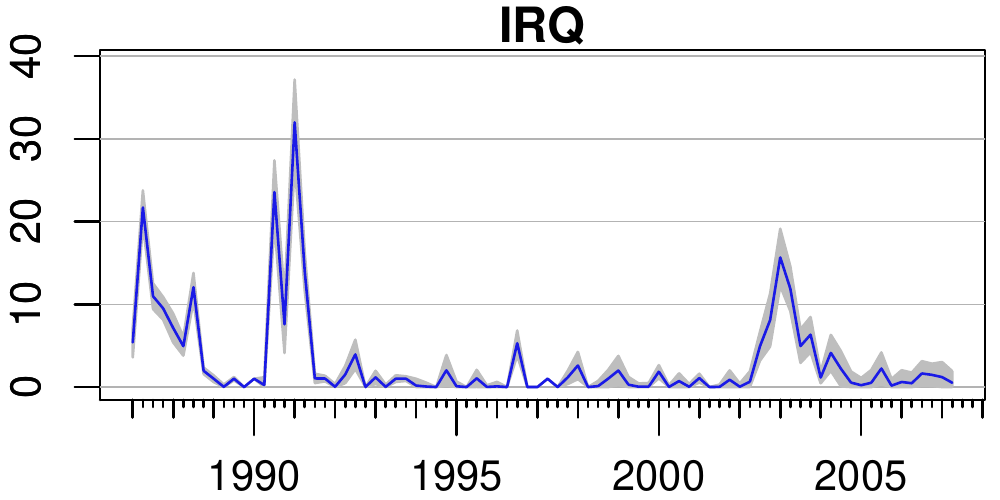}
\tspdf{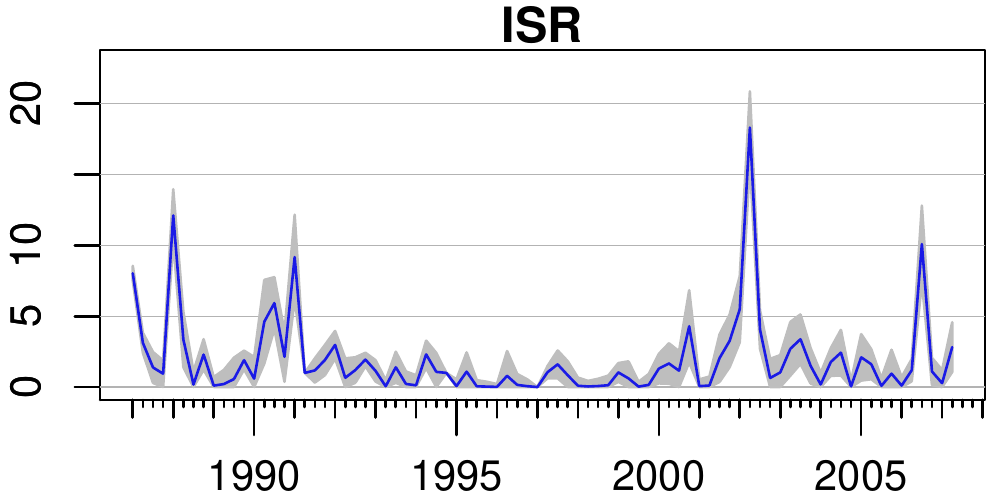}
\tspdf{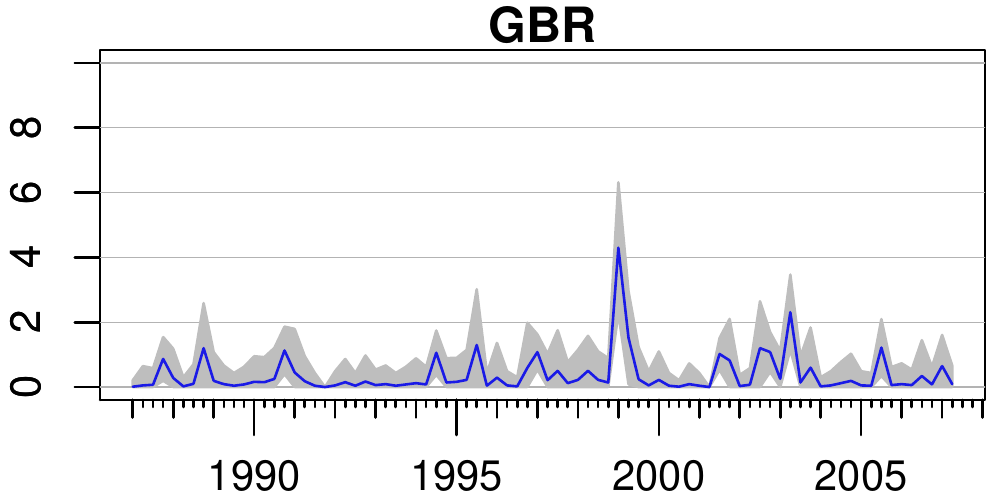}
\tspdf{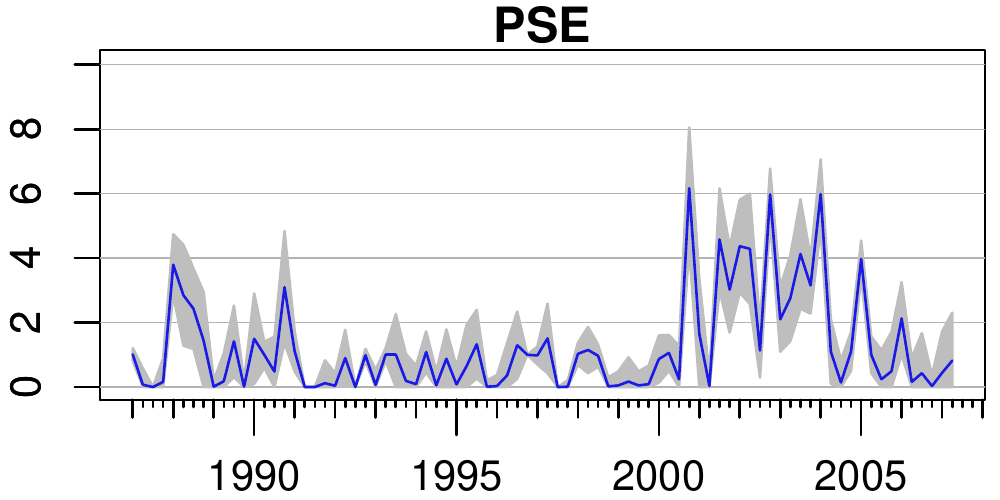}
\tspdf{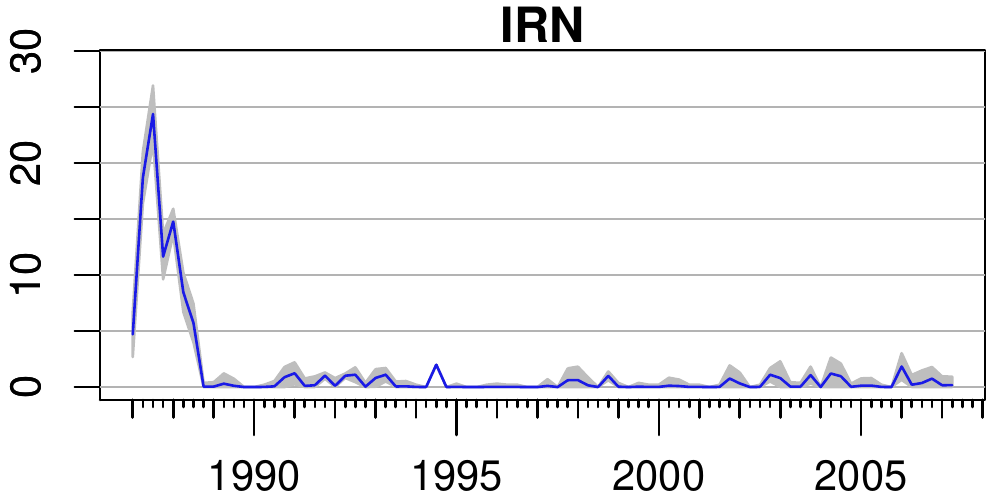}
\tspdf{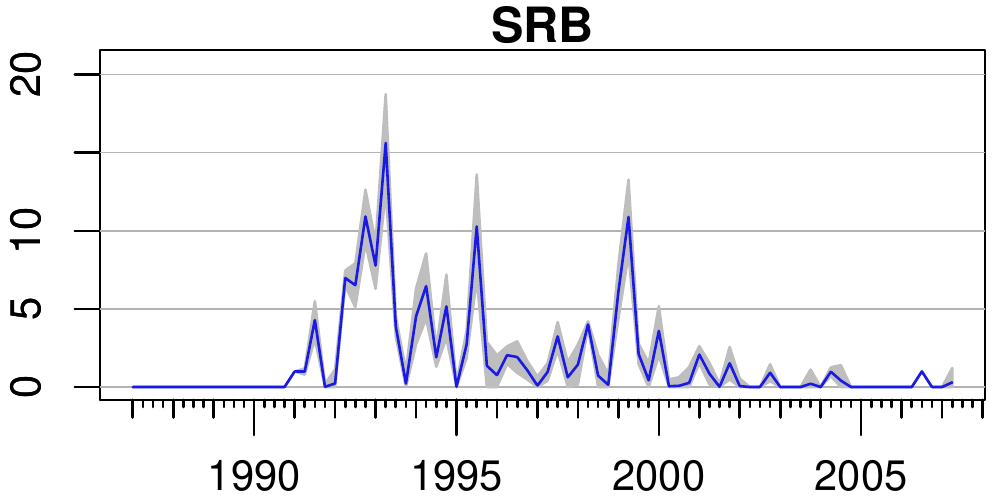}
\tspdf{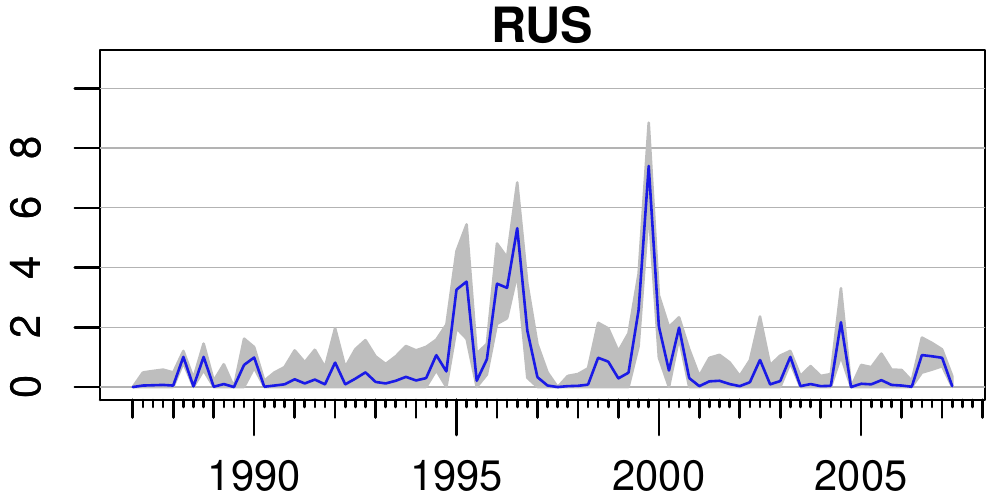}
\tspdf{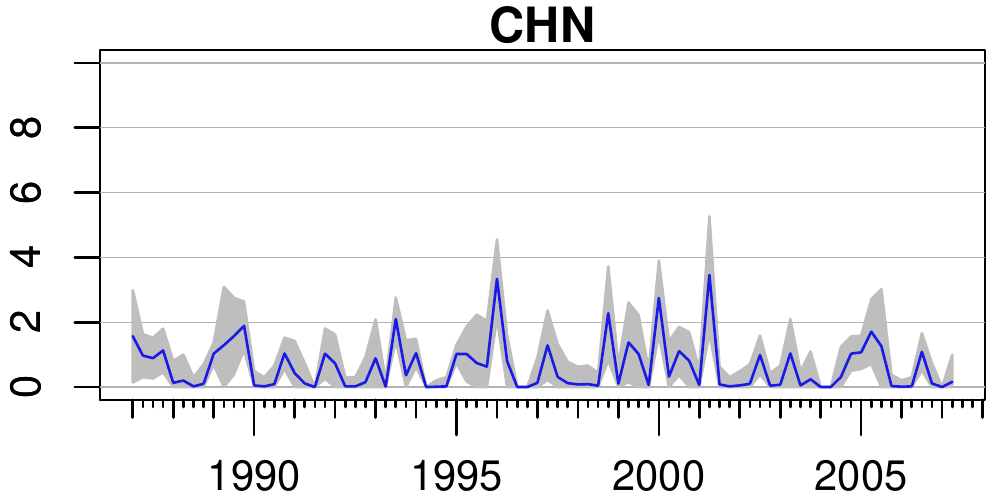}
\tspdf{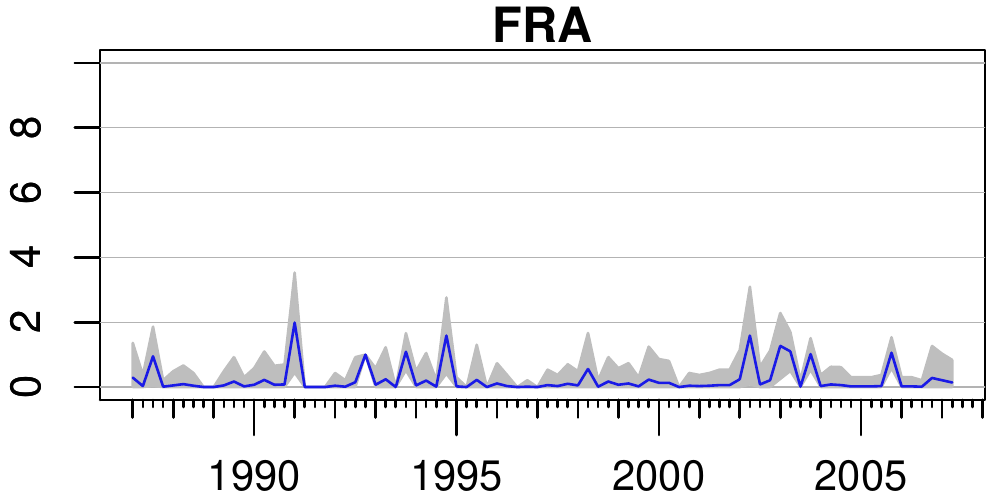}
\tspdf{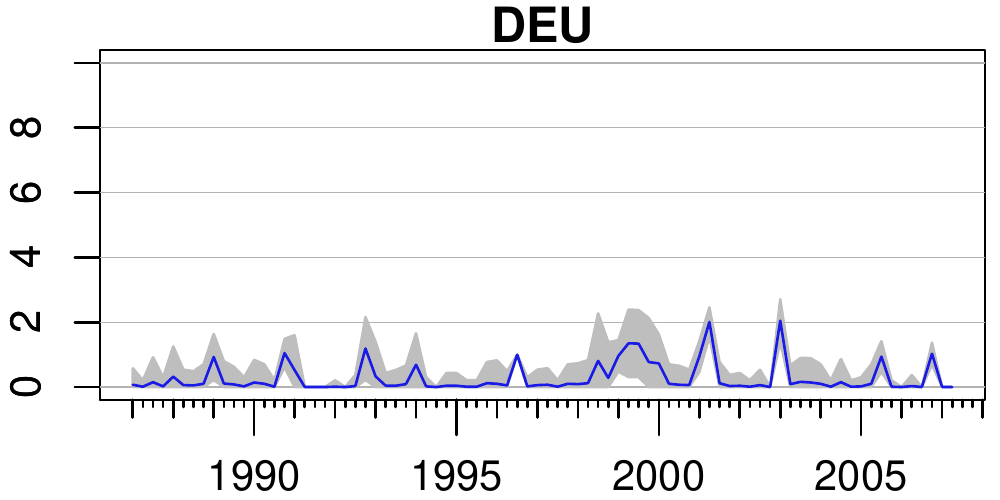}
\tspdf{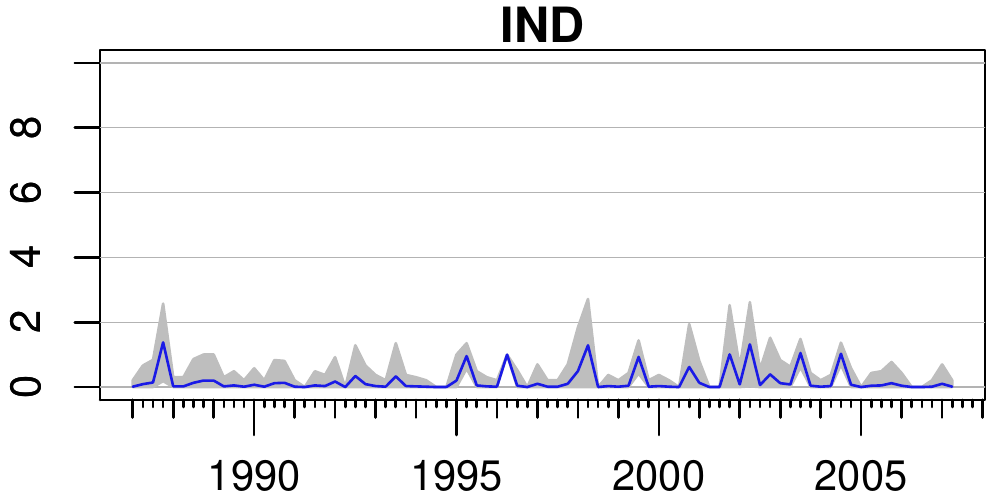}
\tspdf{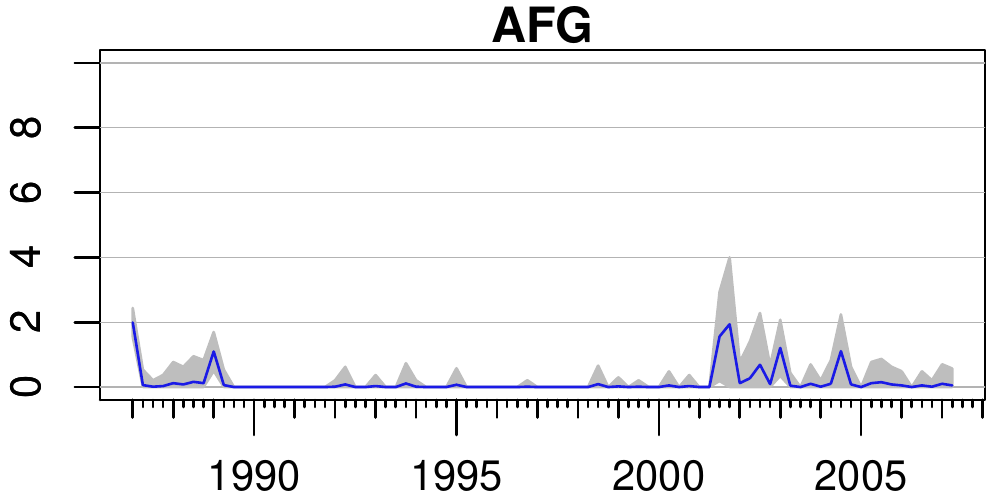}
\tspdf{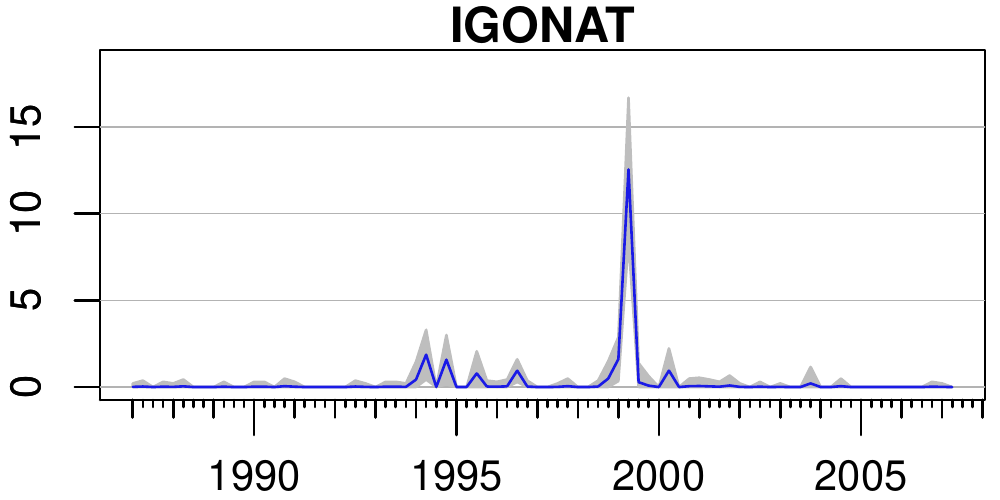}
\tspdf{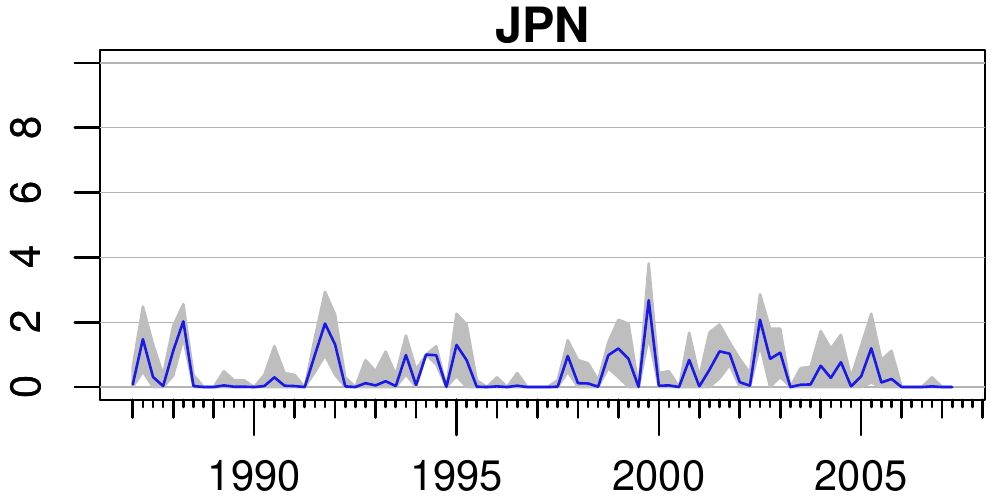}
\tspdf{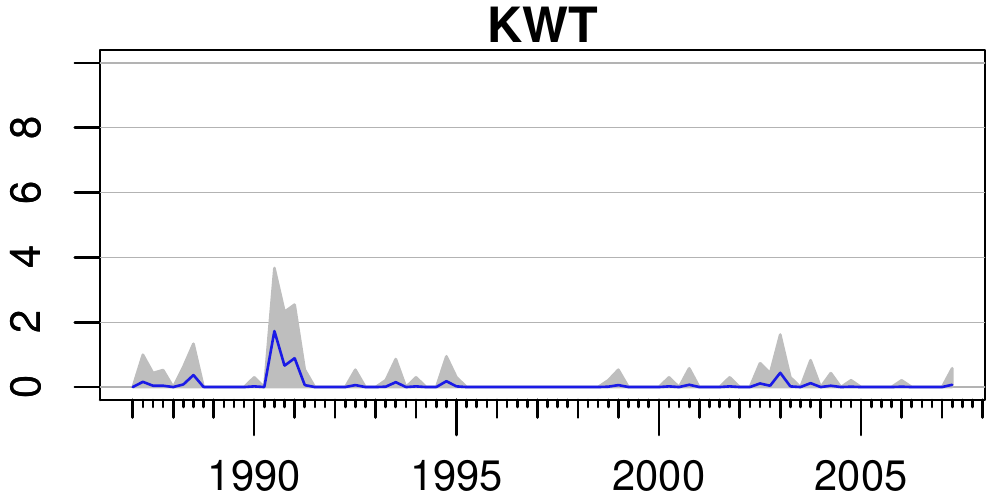}
\tspdf{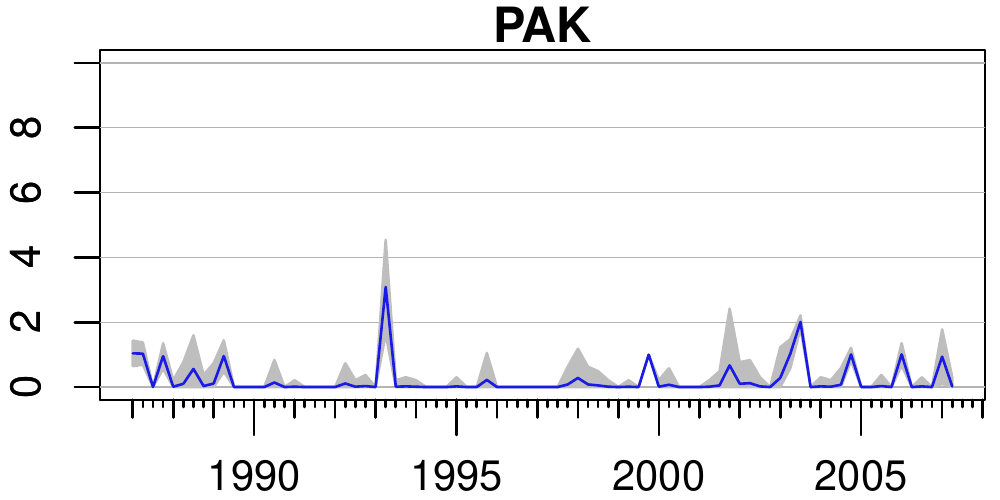}
\tspdf{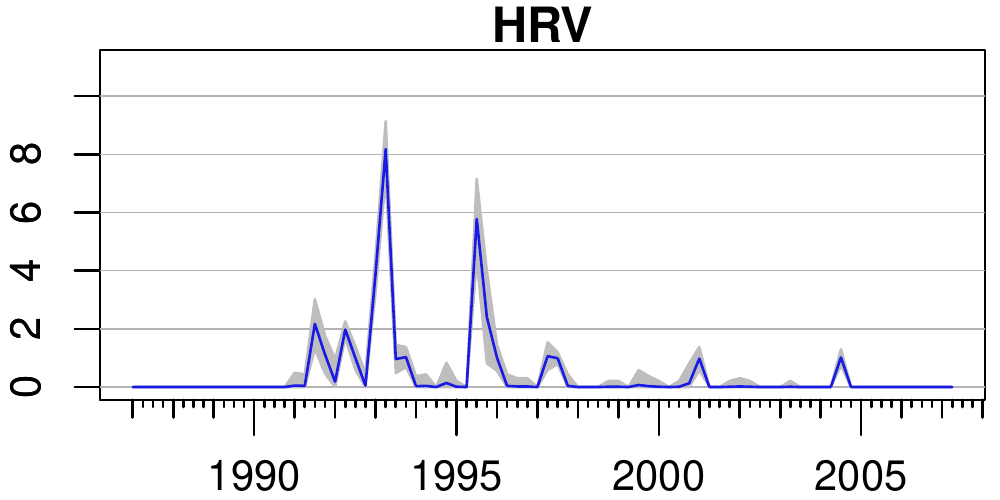}
\tspdf{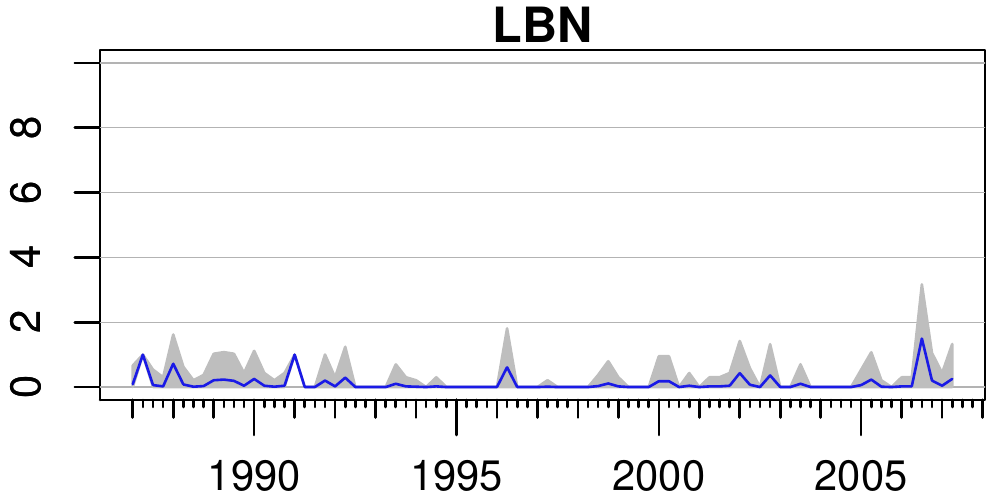}
\tspdf{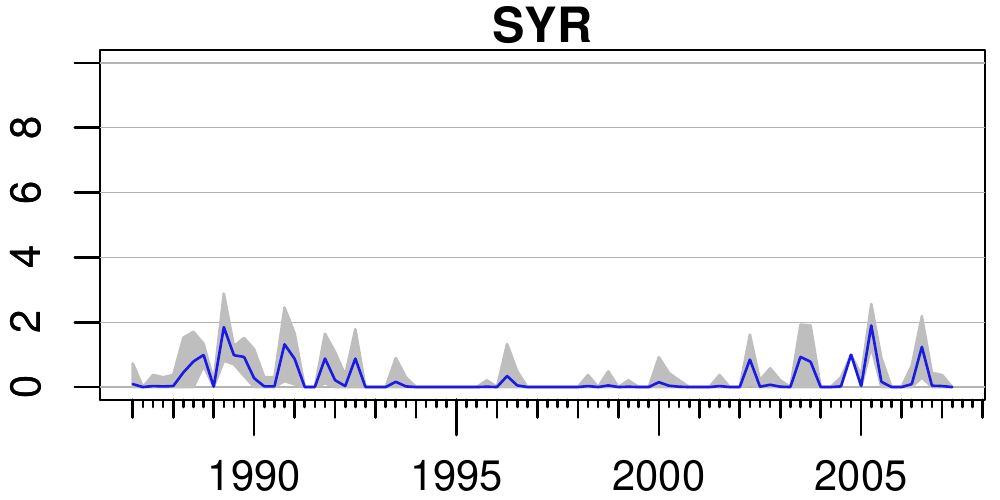}

\end{document}